\documentclass[11pt, a4paper, logo, onecolumn, nonumbering]{googledeepmind}

\pdftrailerid{redacted}

\usepackage{marvosym}
\usepackage{kantlipsum, lipsum}
\usepackage{dsfont}
\usepackage{multirow}
\usepackage{array}
\usepackage{amssymb} %
\usepackage{pifont} %
\usepackage{makecell}
\usepackage{algorithm}
\usepackage{algpseudocode}
\usepackage{amsmath}
\usepackage{multicol}
\usepackage{float}
\usepackage{siunitx}
\usepackage{tabularx}
\usepackage{booktabs}
\usepackage{xspace}

\usepackage{xcolor}
\definecolor{GoogleBlue}{HTML}{1A73E8}
\definecolor{GoogleBlueLight}{HTML}{e8f0fe}
\definecolor{GoogleBlueDark}{HTML}{174ea6}
\definecolor{GoogleRed}{HTML}{D93025}
\definecolor{GoogleGray}{HTML}{767676}
\definecolor{figtextcolor}{HTML}{333333}
\definecolor{perceptioncol}{HTML}{1A73E8}
\definecolor{modelingcol}{HTML}{925BCF}
\definecolor{manipulationcol}{HTML}{C440A7}
\definecolor{reasoningcol}{HTML}{DA2F77}

\usepackage{subcaption}
\usepackage{opensans}
\DeclareCaptionFont{subfigcapsize}{\fontsize{8pt}{8pt}\selectfont}  %
\usepackage{xurl}
\usepackage[colorlinks=true, allcolors=GoogleBlue]{hyperref}
\usepackage[capitalise]{cleveref}
\crefname{section}{Sec.}{Secs.}  %
\crefname{subsection}{Sec.}{Secs.}  %
\crefname{subsubsection}{Sec.}{Secs.}  %
\crefname{appendix}{App.}{Apps.}  %

\usepackage{graphicx}
\graphicspath{{figures/}}

\usepackage{tikz}
\usetikzlibrary{calc}

\usepackage[most]{tcolorbox}

\newcounter{takeaway}
\usepackage{tikz}
\usetikzlibrary{calc}
\newlength{\tikzwidth}
\newlength{\boxpad}
\newlength{\boxgap}
\newlength{\boxtextheight}
\usepackage{xcolor}
\usepackage{tikz}

\usepackage{tcolorbox}
\definecolor{BoxBackground}{RGB}{240, 240, 240} 
\definecolor{BoxFrame}{RGB}{0, 0, 0} 
\definecolor{TitleBackground}{RGB}{0, 0, 0} 
\definecolor{TitleText}{RGB}{255, 255, 255} 
\tcbset{
  academicbox/.style={
    boxsep=5pt,
    left=2pt,
    right=2pt,
    bottom=0.5pt,
    boxrule=0.5pt,
    colback=BoxBackground,
    colframe=BoxFrame,
    colbacktitle=TitleBackground,
    coltitle=TitleText,
    enhanced,
    attach boxed title to top left={yshift=-0.1in,xshift=0.1in},
    boxed title style={boxrule=0pt,colframe=white},
    title={#1},
  }
}
\newtcolorbox{AcademicBox}[1][]{academicbox=#1}

\def\NickName{VINO}

\definecolor{mycolor1}{RGB}{72, 146, 232}
\definecolor{mycolor2}{RGB}{142, 120, 204}
\definecolor{mycolor3}{RGB}{197, 103, 124}

\def\NickNamecolor{\textcolor{mycolor1}{V}\textcolor{mycolor2}{In}\textcolor{mycolor3}{O}}

\title{ \emph{\NickNamecolor}: A Unified \textcolor{mycolor1}{V}isual Generator with \textcolor{mycolor2}{In}terleaved \textcolor{mycolor3}{O}mniModal Context}

\usepackage[numbers, sort&compress]{natbib}
\correspondingauthor{my-email@google.com}
\newcommand{\internshipinfo}{true}

\reportnumber{}

\author[1,\textrm{\dag}]{Junyi Chen}
\author[1]{Tong He}
\author[3]{Zhoujie Fu}
\author[2]{Pengfei Wan}
\author[2]{Kun Gai}
\author[2,\textrm{\Letter}]{Weicai Ye}

\affil[1]{Shanghai Jiao Tong University}
\affil[2]{Kling Team, Kuaishou Technology}
\affil[3]{Nanyang Technology University}

\begin{document}

\begin{abstract}

{\centering
\includegraphics[width=\linewidth]{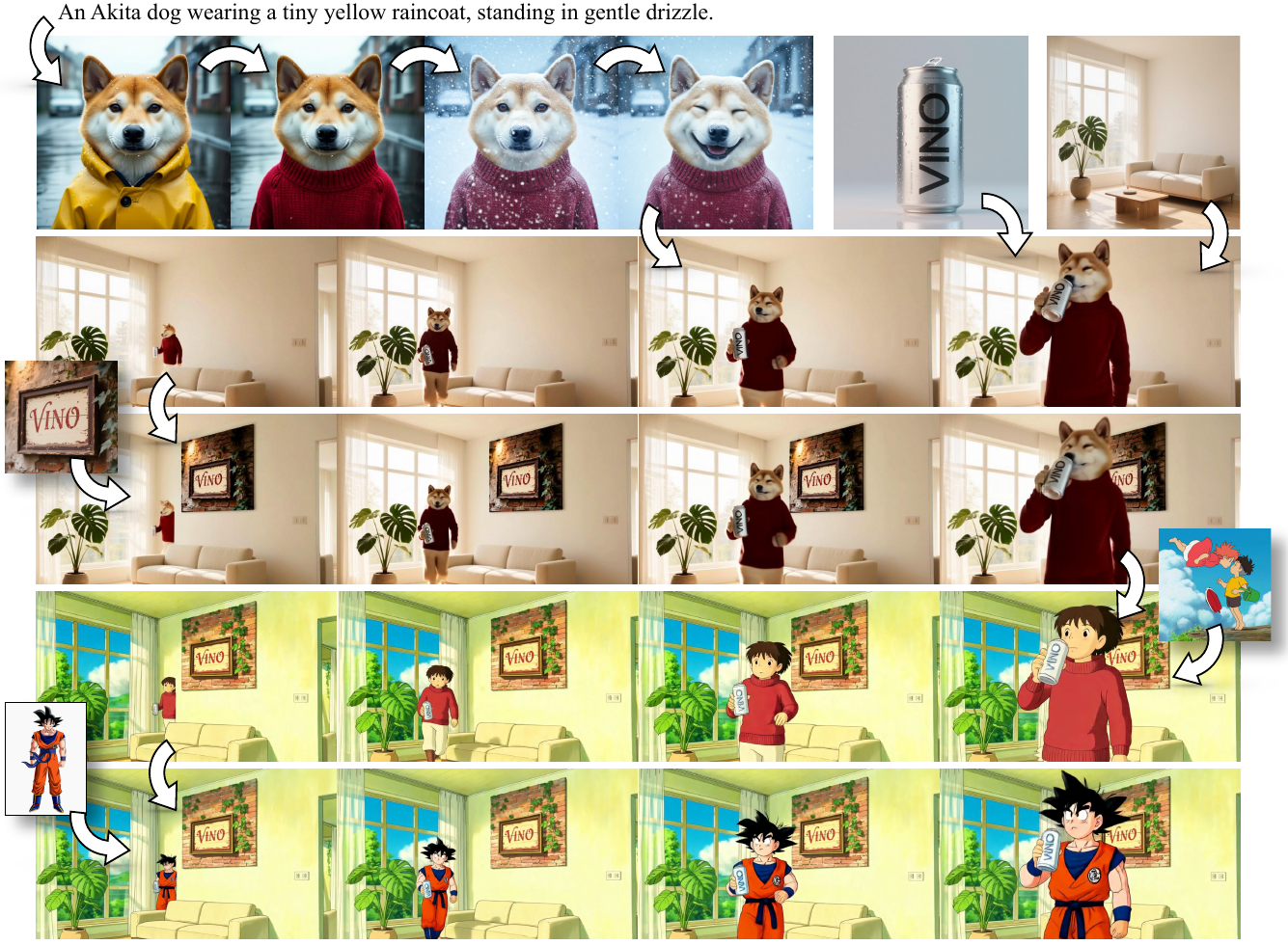}\par}
\vspace{0.7em}

We present \NickName, a unified visual generator that performs image and video generation and editing within a single framework. 
Instead of relying on task-specific models or independent modules for each modality,  \NickName\space uses a shared diffusion backbone that conditions on text, images and videos, enabling a broad range of visual creation and editing tasks under one model.
Specifically, \NickName\space couples a vision–language model (VLM) with a Multimodal Diffusion Transformer (MMDiT), 
where multimodal inputs are encoded as interleaved conditioning tokens, and then used to guide the diffusion process. 
This design supports multi-reference grounding, long-form instruction following, and coherent identity preservation across static and dynamic content, while avoiding modality-specific architectural components. 
To train such a unified system,
we introduce a multi-stage training pipeline that progressively expands a video generation base model into a unified, multi-task generator capable of both image and video input and output.
Across diverse generation and editing benchmarks, 
\NickName\space demonstrates strong visual quality, faithful instruction following, improved reference and attribute preservation, and more controllable multi-identity edits.
Our results highlight a practical path toward scalable unified visual generation, and the promise of interleaved, in-context computation as a foundation for general-purpose visual creation. 

\vspace{1em}
{\includegraphics[height=1em]{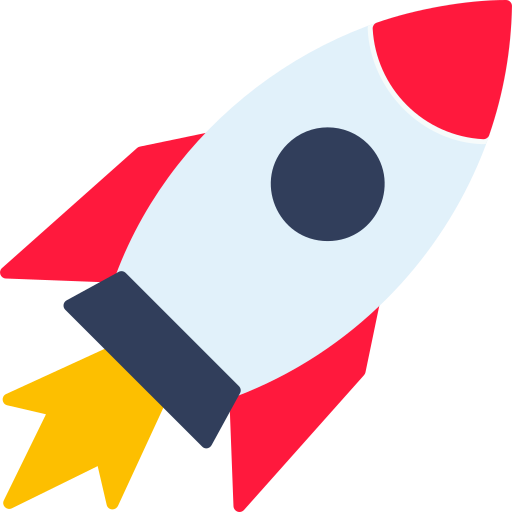}} Project Page: \url{https://sotamak1r.github.io/VINO-web/} \\
{\includegraphics[height=1em]{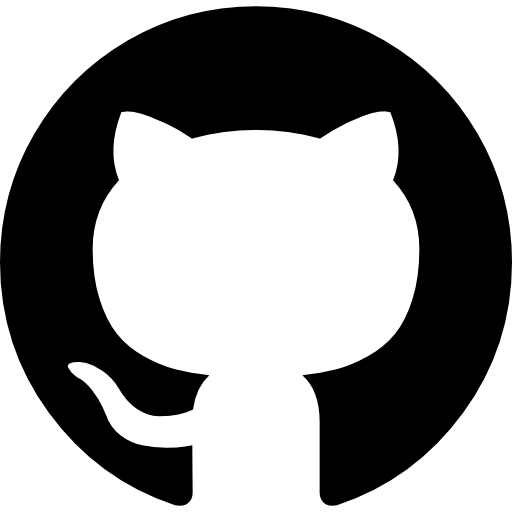}} Github Code: \url{https://github.com/SOTAMak1r/VINO-code/} \\

\end{abstract}

\newpage
\maketitle

\begin{figure}
    \centering
    \includegraphics[width=\linewidth]{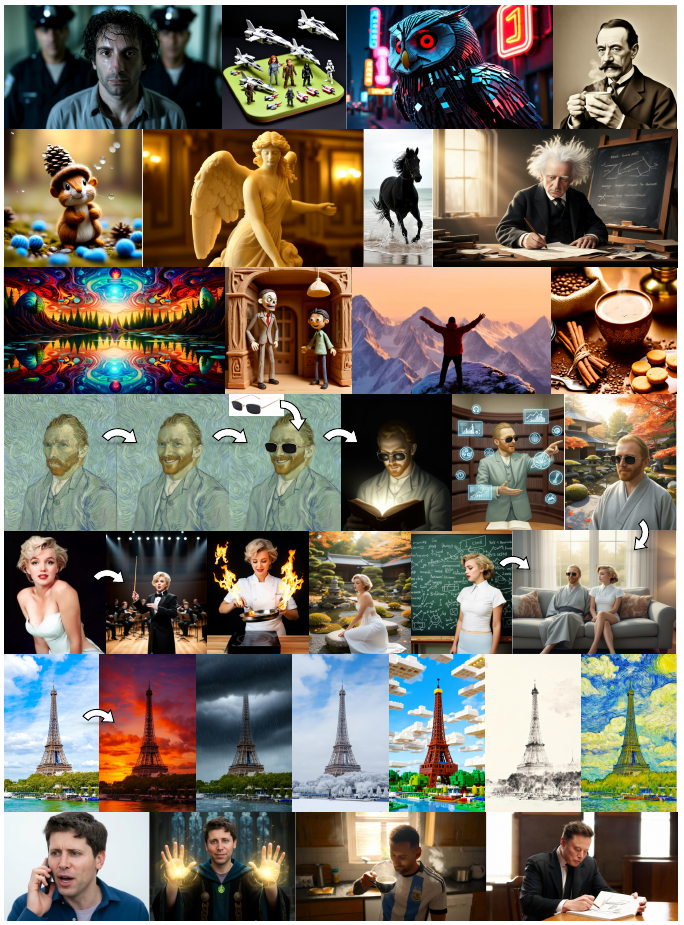}
    \caption{Showcase of \NickName\space in image generation and image editing.}
    \label{fig:teaser_image}
\end{figure}

\begin{figure}
    \centering
    \includegraphics[width=\linewidth]{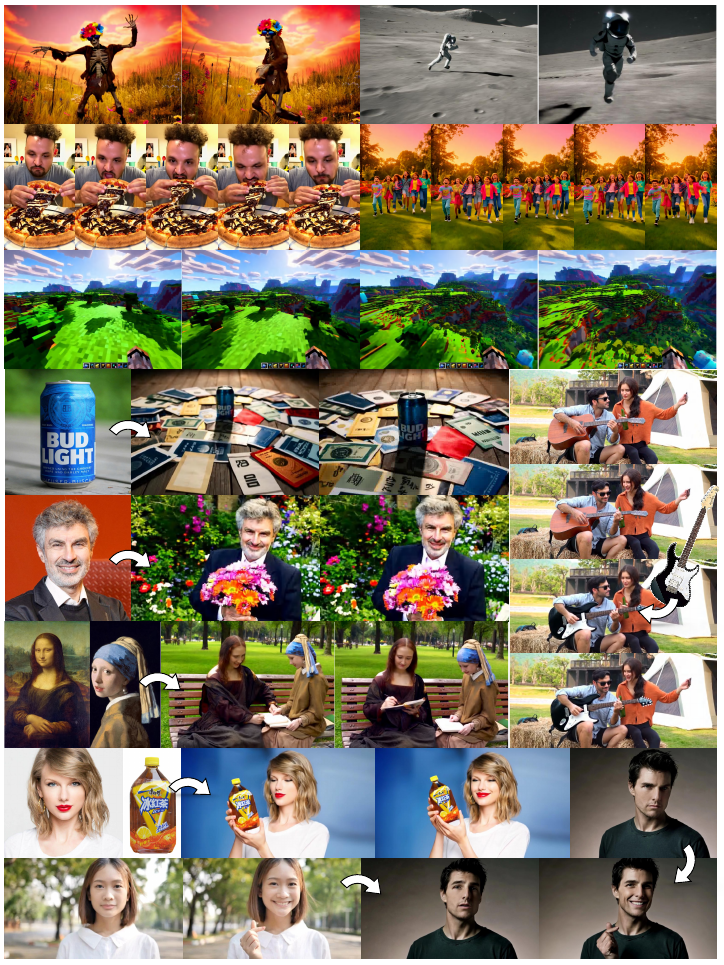}
    \caption{Showcase of \NickName\space in video generation and video editing.}
    \label{fig:teaser_video}
\end{figure}

\newpage

\section{Introduction}
\label{sec:intro}

Recent diffusion models \citep{flow,normalflow,flowmatching,ddpm,sgm} achieve high-fidelity image and video synthesis, and multimodal assistants \citep{gpt4v,qwen3vl,llava,videollama} show that unified perception across text and vision is feasible.
Yet, visual generation pipelines in practice remain fragmented: text-to-image  \citep{sd3,flux}, text-to-video \citep{kling,sora,veo3}, and visual editing models \citep{step1xedit,qwenimage,vace} are developed and deployed separately, 
while multimodal LLMs offer unified perception but still rely on external diffusion backbones or decoders for high-resolution visual generation \citep{metaquery,blip3o}.
This motivates the goal of building a unified visual generation framework.

Building such a unified generator, however, poses two fundamental challenges.
First, generation tasks often rely on rich, descriptive captions \citep{dalle3}, 
whereas editing tasks use short, imperative instructions that modify localized regions or attributes.
Second, when text, images, videos, and other guidance signals are provided simultaneously, current models lack mechanisms to reliably disentangle and prioritize multimodal signals, resulting in semantic conflicts or inconsistent conditioning effects.
These challenges suggest that a unified generator should process multimodal cues jointly and reason about their interactions.

To address these limitations, we propose \NickName, a unified visual generator that performs interleaved generation and editing across both images and videos. 
\NickName\space couples a vision–language model (VLM) with a Multimodal Diffusion Transformer (MMDiT), where the VLM processes all control signals (text, reference images and video) and exposes them as unified conditioning tokens, while the MMDiT operates as a token-based diffusion backbone over latent images and videos.
This formulation enables a single diffusion backbone to accommodate multimodal control sources without task-specific modules.

Beyond unifying control signals, we introduce two architectural components:
first, a key component is the use of learnable query tokens at the VLM input. 
Inspired by recent work \citep{metaquery}, these learnable tokens provide a flexible interface between high-level instructions and low-level diffusion features and are jointly optimized with the generator. 
We integrate the learnable tokens processed by the VLM together with tokens from all modalities into the MMDiT. 
Empirically, this formulation yields smoother optimization, improved stability, and stronger generation and editing quality.
Second, to preserve detailed reference information, we not only use the VLM-extracted features as conditioning signals, but also injects the corresponding VAE latents into the MMDiT \citep{bagel,showo2,mogao}.
Crucially, for each reference image or video, we reuse the same paired start–end tokens that wrap its VLM tokens to additionally wrap its VAE latent tokens. 
In this way, both the semantic (VLM) and latent (VAE) representations of the same reference are associated with the same boundary markers, enabling the MMDiT to consistently recognize and group features originating from the same visual source.

While the architecture provides unified conditioning, directly training such a model remains challenging due to heterogeneous instruction formats.
First, we utilize image and video data with long captions to align the output space of the VLM with the condition space of MMDiT. 
To adapt the model to short captions and prompt the learnable tokens to acquire the ability to refine short captions, we introduce a mixed training phase that incorporates both long and short captions. 
After the model has adapted to both long and short captions, we proceed to train its multi-task capabilities.

We evaluate \NickName\space across diverse image and video generation and editing tasks, 
including text-to-image, text-to-video, instruction-based image editing, reference-driven video generation, and instruction-based video editing. 
Empirically, \NickName\space delivers strong visual quality, improved identity preservation and faithful instruction following across both images and videos, 
providing a more scalable path toward unified visual generation than existing task-specific diffusion pipelines.

In summary, this paper makes the following contributions:

\begin{enumerate}
    \item We present \NickName, 
    a unified diffusion-based architecture that handles both image and video generation and editing within a single framework by coupling a VLM with a MMDiT.

    \item We incorporate learnable query tokens into the VLM's input and provide an empirical study demonstrating 
    that they improve multimodal conditioning, stabilize optimization.

    \item We propose a token-boundary mechanism that reuses special VLM tokens within the MMDiT's VAE latent stream to maintain consistent grounding across semantic and latent representations of reference images and videos.
    This reduces identity swapping and attribute leakage in complex, multimodal reference scenes.

    \item We introduce a progressive training strategy that upgrades a pretrained video model into a multi-task visual generator while preserving its original visual generation strength.

\end{enumerate}

\begin{figure}[h]
    \centering
    \includegraphics[width=\linewidth]{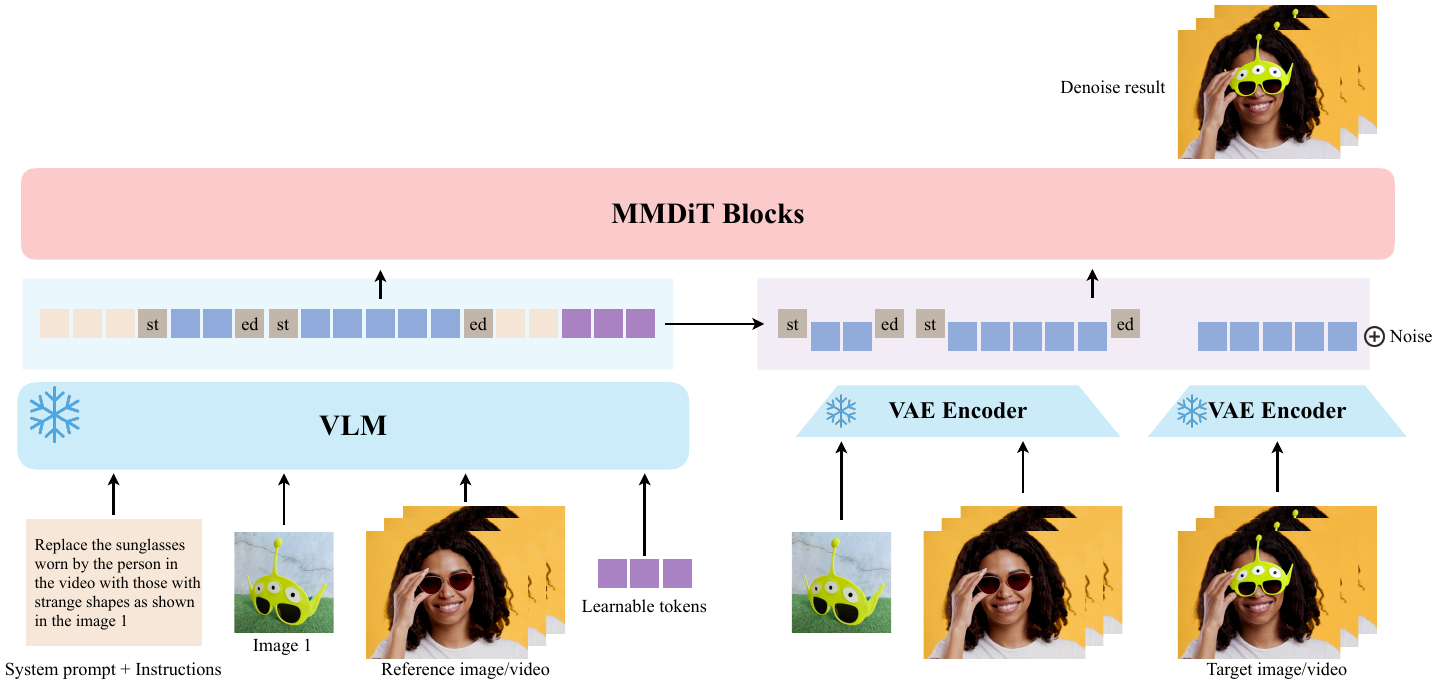}
    \caption{\textbf{Overview of the \NickName\space pipeline.}
Our unified framework conditions generation on an interleaved omnimodal context that jointly encodes system prompts, prompts/instructions, reference images/videos, and learnable tokens. 
A frozen VLM processes textual instructions together with visual references, producing multimodal embeddings that are augmented with learnable tokens (purple) and separated by special tokens (vision start token \includegraphics[height=0.9em]{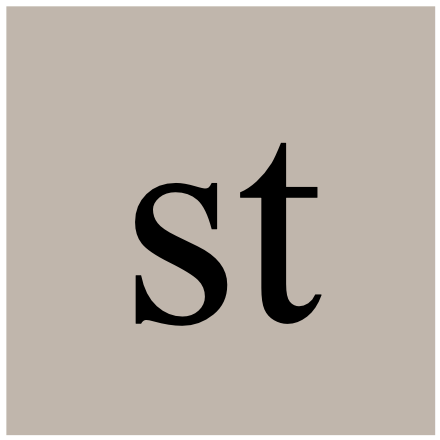} and vision end token \includegraphics[height=0.9em]{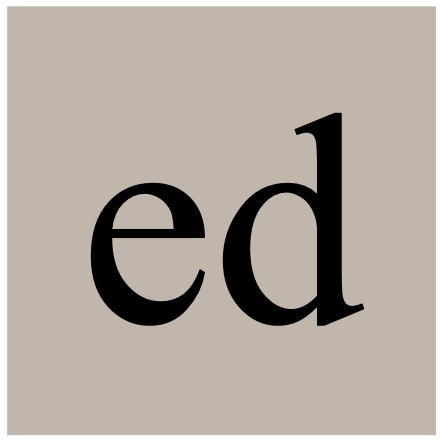}). 
These interleaved multimodal representations are fed into the MMDiT blocks, which also receive VAE latents from the reference images or video. 
The MMDiT model performs denoising conditioned on the full multimodal context, enabling \NickName\space to execute image and video generation as well as instruction-based editing within a single unified architecture.
}
    \label{fig:pipeline}
\end{figure}

\section{Methods}
\label{sec:methods}

In this section, we present our unified framework for multi-modal image and video generation/editing. 
Our objective is to design a system that accepts heterogeneous control signals—textual instructions, reference images or videos, and learnable tokens—and uses them to guide a diffusion-based visual generator. 
Following the high-level model pipeline (Figure. \ref{fig:pipeline}), we structure this section around three central components. 
We first describe how multi-modal conditions are processed by the vision–language model (VLM) to obtain coherent feature representations in Section \ref{sec:Multi-Modal Conditions}. 
We then explain how these encoded conditions are injected into the Multi-Modal Diffusion Transformer (MMDiT) without causing ambiguity or incorrect cross-modal grounding in Section \ref{sec:Interleaved OmniModal Context}. 
Finally, we detail the training strategy that makes the entire architecture a unified, multi-task visual generator capable of supporting a wide spectrum of editing and generation tasks in Section \ref{sec:Training the Unified Multi-Task Visual Generator}.

\subsection{Multi-Modal Conditions}
\label{sec:Multi-Modal Conditions}

\begin{figure}[t]
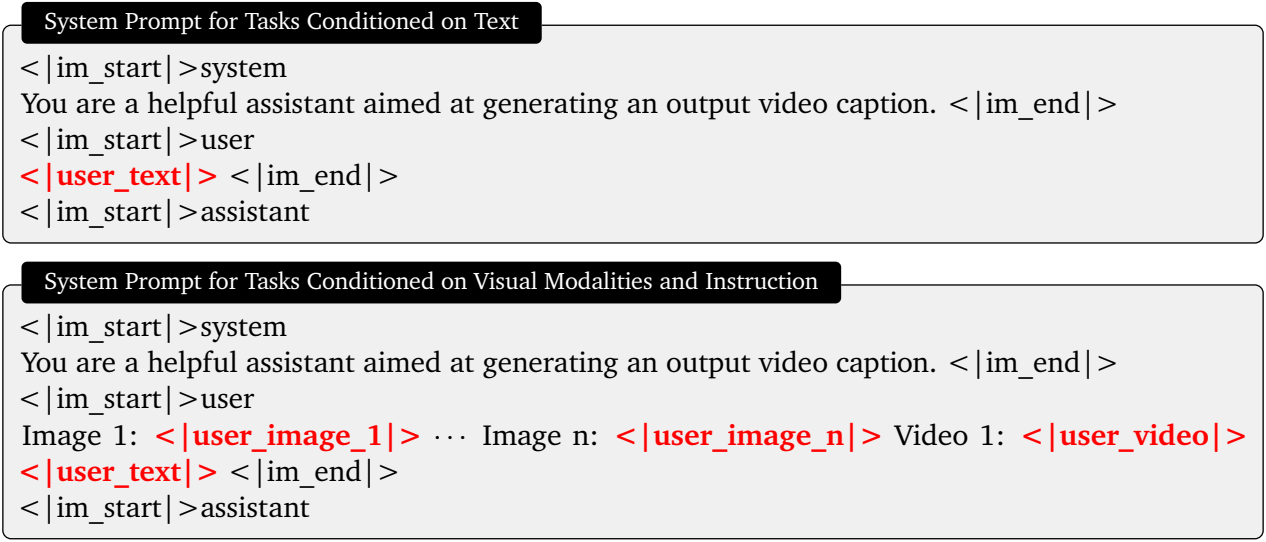


\begin{AcademicBox}[\footnotesize System Prompt for Tasks Conditioned on Text]
<|im\_start|>system \\ 
You are a helpful assistant aimed at generating an output video caption.
<|im\_end|> \\
<|im\_start|>user \\
\textbf{\textcolor{red}{<|user\_text|>}}
<|im\_end|> \\
<|im\_start|>assistant
\end{AcademicBox}

\begin{AcademicBox}[\footnotesize System Prompt for Tasks Conditioned on Visual Modalities and Instruction]
<|im\_start|>system \\ 
You are a helpful assistant aimed at generating an output video caption.
<|im\_end|> \\
<|im\_start|>user \\
Image 1: \textbf{\textcolor{red}{<|user\_image\_1|>}} $\cdots$  Image n: \textbf{\textcolor{red}{<|user\_image\_n|>}}
Video 1: \textbf{\textcolor{red}{<|user\_video|>}}
\textbf{\textcolor{red}{<|user\_text|>}}
<|im\_end|> \\
<|im\_start|>assistant
\end{AcademicBox}

\caption{
System prompt for different condition task, where \textbf{\textcolor{red}{<|user\_text|>}}, \textbf{\textcolor{red}{<|user\_image|>}} and \textbf{\textcolor{red}{<|user\_video|>}} denote the user-provided input conditions across different modalities.
For brevity, we omit the \textbf{<|vision\_start|>} and \textbf{<|vision\_end|>} tokens for the visual modalities.
}
\label{fig:sys_prompt}
\end{figure}

To handle diverse forms of input, we employ a frozen VLM model as the front-end encoder for all language and visual conditions. 
As illustrated in the Figure~\ref{fig:sys_prompt}, the system prompt varies with the presence and number of input modalities. 
When no visual modality is provided, the user supplies only a text input, which serves as the sole condition for text-to-image or text-to-video generation. 
When visual inputs are present, they are first sorted by type (image, then video) and placed at the beginning of the prompt, each assigned a unique identifier such as Image 1 or Video 1. 
The user can then reference these identifiers in the text input to specify different visual conditions, enabling complex multimodal control. 
In addition, we append a set of learnable tokens at the end of the prompt to extract cross-modal features into a shared space. 
These tokens are also handled using causal masking rather than granting full bidirectional attention \citep{metaquery}.
Finally, we use the penultimate layer hidden states of the VLM as the encoded conditioning, apply a two-layer Multi-Layer Perceptron (MLP) for feature projection, and then feed them into the subsequent MMDiT.

\subsection{Interleaved OmniModal Context}
\label{sec:Interleaved OmniModal Context}

\begin{figure}
    \centering
    \includegraphics[width=\linewidth]{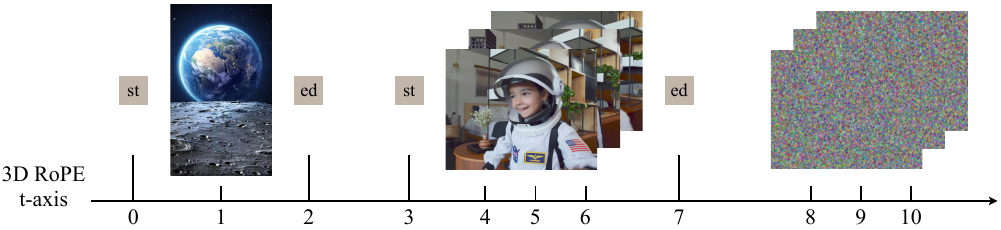}
    \caption{\textbf{3D RoPE strategy for the VAE branch in \NickName.}
We apply a unified 3D RoPE schedule along the temporal axis to interleave different visual modalities in the MMDiT VAE branch. 
Each modality—single reference images, multi-frame reference videos, and noisy target latents—is placed on a shared RoPE timeline, separated by special tokens, which are projected from the VLM output. 
This structured RoPE layout enables the model to distinguish heterogeneous visual sources.
}
    \label{fig:rope}
\end{figure}

Although the VLM provides robust high-level multimodal semantics, it significantly compresses visual information, resulting in a lack of fine-grained spatial details and texture fidelity.
Consequently, it cannot adequately handle tasks requiring precise structural control, such as local editing.
To compensate for this information bottleneck, 
we complement the VLM embeddings with VAE-encoded latents of all visual modalities. 
As illustrated in the Figure~\ref{fig:rope}, these VAE latents are arranged following the same ordering used in the VLM, and place the noised image/video latents at the end. 
However, simply concatenating image and video latents introduces ambiguity. 
To uniquely distinguish different visual conditions and to align each VAE latent with its corresponding VLM feature, we reuse the VLM’s \textbf{<|vision\_start|>} and \textbf{<|vision\_end|>} embeddings. 
After projecting these embeddings through an MLP to match the MMDiT input dimension, they are used to mark the boundaries of each visual latent block. 
This explicit boundary marking serves as a strong positional cue, allowing the attention mechanism to correctly effectively distinguish and interpret distinct visual conditioning inputs within the sequence.

\subsection{Training the Unified Multi-Task Visual Generator}
\label{sec:Training the Unified Multi-Task Visual Generator}

\begin{figure}[h]
    \centering
    \includegraphics[width=\linewidth]{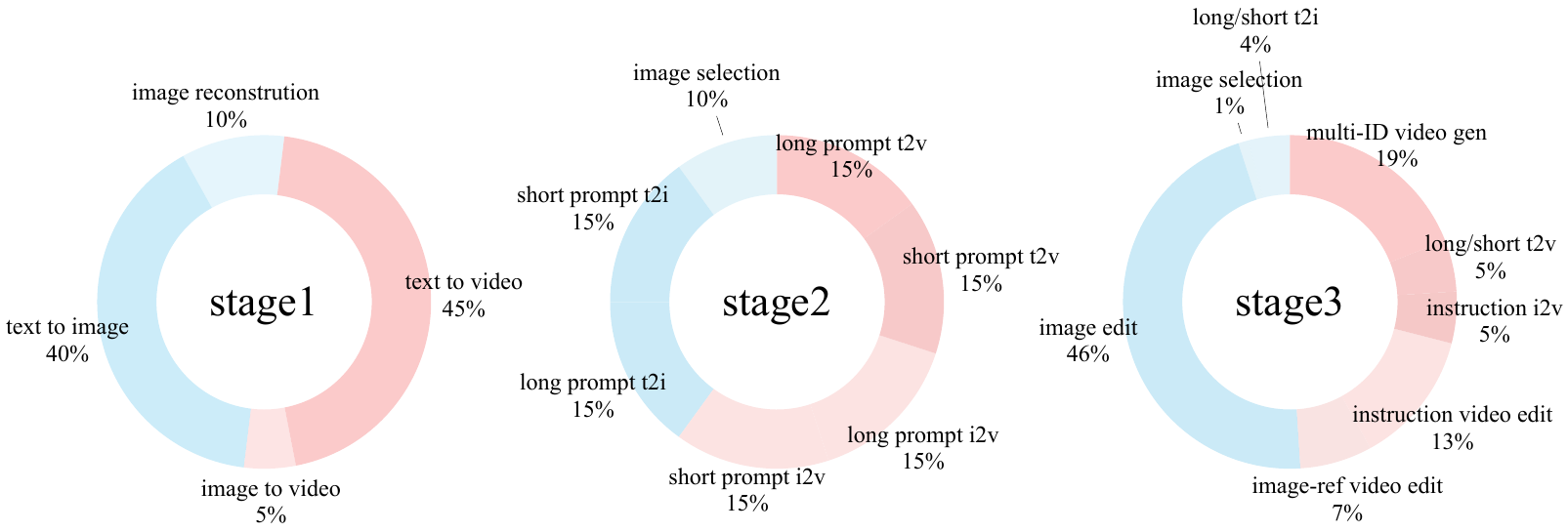}
    \caption{\textbf{Training data distribution across the stages.}
This progressive strategy gradually transforms the base model from a pure text-to-video generator into a capable multi-task visual generator.
}
    \label{fig:data}
\end{figure}

To build a unified visual generator that supports multimodal conditioning, we start from a text-to-video diffusion model, as it already provides strong priors for temporal dynamics. 
To replace the original text encoder, we first align the output space of VLM with the model’s native text encoder. 
In this initial stage, only a two-layer MLP connector is trained to map between the two embedding spaces.
Modern text-to-video models typically rely on long, well-structured textual prompts, whereas editing tasks often involve short instructions, creating a distributional gap. 
To bridge this gap, we adopt a progressive training strategy that gradually shifts the input-condition distribution. 
Specifically, we treat short prompts as an intermediate form between long prompts and concise editing instructions. 
In the second stage, we train the model with a mixture of long and short prompts to ensure robustness across both forms, and we begin updating the MMDiT parameters during this stage.
Once the model has adapted to short-prompt inputs, we enter the final stage, where full multi-task mixed training is performed. 
The data mixture ratio for each stage is illustrated in the Figure~\ref{fig:data}. 
This allows the model to smoothly transition from structured text-to-video conditioning to instruction-based multimodal generation and editing.

\section{Experiments}
\label{sec:experiments}

In this section, we present experimental results demonstrating \NickName’s performance across a variety of visual generation and editing tasks. 
We compare against several advanced baselines, including both open-source and close-source systems. 
Notably, most competing methods remain proprietary and do not offer a unified generator comparable to \NickName. 
Additional visualizations are available on our project webpage.

\begin{table}[ht]
    \centering
    \caption{Training hyperparameters and trainable modules for the three-stage progressive training pipeline.}
    \label{tab:training_stages}
    \sisetup{group-digits = false} %
    \begin{tabular}{l c c c} 
        \toprule
          & \textbf{Stage 1} & \textbf{Stage 2} & \textbf{Stage3} \\
        \midrule
        Learning rate & $1 \times 10^{-4}$  & $4 \times 10^{-5}$  & $2 \times 10^{-5}$   \\
        LR scheduler &  Constant & Constant  & Constant \\
        Weight decay & 0.01 & 0.01 & 0.01 \\
        Gradient norm clip  & 1.0 &  1.0 &  1.0 \\
        Training steps & 20k & 4k & 16k \\
        Training samples & $\mathcal{O}(100)k$ & $\mathcal{O}(100)k$ & $\mathcal{O}(100)k$ \\
        AdamW betas & $(0.9, 0.99)$ & $(0.9, 0.95)$ & $(0.9, 0.95)$ \\
        Use ema & no & yes & yes \\
        Ema decay & - & $0.9999$ & $0.9999$ \\             
        Frame area & $320^2$ & $640^2$ & $640^2$ \\
        Timestep shift & $3$ & $5$ & $5$ \\     
        \midrule
        Trainable modules & \makecell{connectors\\learnable tokens\\ \mbox{}}  & \makecell{connectors\\learnable tokens\\MMDiT} & \makecell{connectors\\learnable tokens\\MMDiT}\\

        \bottomrule
    \end{tabular}
    
\end{table}

\subsection{Implement Details}
\label{sec:Implement Details}

\paragraph{Datasets Composition.}
As illustrated in Figure \ref{fig:data}, 
our training pipeline allows for a progressive capability acquisition, organized into multiple stages where each targets a distinct subset of tasks.
To support this staged paradigm, 
we curate a comprehensive mixture of datasets, combining large-scale open-source image/video collections
\citep{openvid1m,laion5b,picobanana400k,nanobanana150k,ditto,omniworld,camclonemaster,recammaster,miradata,blip3o,opens2v,panda70m} 
and high-quality distillation data from open-source models
\citep{vace,wan,wananimate,deepverse,iclight}, 
ensuring broad coverage across visual generation and visual editing. 
To handle this diverse data efficiently, we implement a dynamic resolution bucketing strategy. 
Instead of resizing all inputs to a fixed square resolution, we dynamically rescale images and videos based on their original dimensions.
Specifically, within each training stage, we ensure that all samples have similar effective areas (i.e., total pixel count) while preserving their original aspect ratios. 
This strategy maintains a balanced computational load across GPUs at every training step, while simultaneously enabling the model to learn and generalize across a range of spatial resolutions.

\paragraph{Base Models.}
Regarding the architecture, we integrate strong pre-trained priors to accelerate convergence.
We adopt Qwen3VL-4B-Instruction \citep{qwen3vl}, one of the most recent and advanced vision–language models, as our multimodal encoder to provide 
strong understanding capabilities. 
For the visual generator, we initialize our model with HunyuanVideo \citep{hunyuanvideo}, an advanced text-to-video backbone. 
As shown in Table \ref{tab:geneval}, compared to other open-source baselines,  HunyuanVideo demonstrates better prompt adherence, producing visually coherent outputs even with concise or minimal textual descriptions, making it an ideal foundation for our instruction-driven generation tasks.

\paragraph{Training details.}
Table \ref{tab:training_stages} provides a comprehensive overview of the hyperparameters employed across our three-stage training pipeline.
This progressive curriculum is engineered to incrementally expand the model’s capabilities while ensuring stability and efficient optimization.
For all stages, we train using DeepSpeed ZeRO-2 \citep{zero}, which reduces memory consumption and enables training large modules efficiently. 
We further apply gradient checkpointing to the MMDiT backbone to reduce activation memory during video-heavy tasks. 
To ensure that each GPU can sustain a local batch size of 1, we dynamically adjust the number of video frames and number of reference images depending on the task. 
Since different tasks have different computational costs, 
we enforce strict task synchronization across all GPUs at each step, preventing workload imbalance and ensuring optimal training throughput.


\begin{table}[ht]
    \centering
    \caption{\textbf{Geneval \citep{geneval} results comparing \NickName\space with image and video models.}
Compared with its base model \citep{hunyuanvideo}, \NickName\space maintains strong text-to-image ability using only a small proportion of T2I samples in Stage 3. ${\dag}$ refer to the methods using LLM rewriter \citep{bagel}.
}
    \label{tab:geneval}
    \sisetup{group-digits = false} 
    
    \resizebox{\linewidth}{!}{
        \begin{tabular}{l l c c c c c c c} 
            \toprule
            \textbf{Type} & \textbf{Name} & \textbf{Single obj.} & \textbf{Two obj.} & \textbf{Counting} & \textbf{Colors} & \textbf{Position} & \textbf{Color attr.} & \textbf{Overall $\uparrow$} \\
            \midrule
            \multirow{8}{*}{ \rotatebox[origin=c]{90}{Image Models} }
            & SDv2.1 \citep{ldm} & 0.98 & 0.51 & 0.44 & 0.85 & 0.07 & 0.17 & 0.50 \\
            & SDxl \citep{sdxl}  & 0.98 & 0.74 & 0.39 & 0.85 & 0.15 & 0.23 & 0.55 \\
            & Sana \citep{sana} & 0.99 & 0.77 & 0.62 & 0.88 & 0.21 & 0.47 & 0.66 \\
            & Emu3-Gen \citep{emu3} & 0.99 & 0.81 & 0.42 & 0.80 & 0.49 & 0.45 & 0.66 \\
            & DALLE 3 \citep{dalle3} & 0.96 & 0.87 & 0.47 & 0.83 & 0.43 & 0.45 & 0.67 \\
            & FLUX.1-dev \citep{flux} & 0.99 & 0.81 & 0.79 & 0.74 & 0.20 & 0.47 & 0.67 \\
            & SD3 \citep{sd3}    & 0.99 & 0.94 & 0.72 & 0.89 & 0.33 & 0.60 & 0.74 \\
            & Playgroundv3 \citep{omnigen} & 0.99 & 0.95 & 0.72 & 0.82 & 0.50 & 0.54 & 0.76 \\
            \midrule
            \multirow{2}{*}{ \rotatebox[origin=c]{90}{Video Models} }
            & Wan2.1-14B \citep{wan}                     & 0.88 & 0.55 & 0.51 & 0.71 & 0.16 & 0.25 & 0.51 \\
            & HunyuanVideo \citep{hunyuanvideo}          & 0.95 & 0.77 & 0.34 & 0.77 & 0.43 & 0.44 & 0.61 \\
            & HunyuanVideo$^{\dag}$ \citep{hunyuanvideo} & 0.96 & 0.88 & 0.43 & 0.84 & 0.66 & 0.47 & 0.71 \\
            & \cellcolor{gray!20}\NickName                                  & \cellcolor{gray!20}0.95 & \cellcolor{gray!20}0.72 & \cellcolor{gray!20}0.33 & \cellcolor{gray!20}0.80 & \cellcolor{gray!20}0.21 & \cellcolor{gray!20}0.49 & \cellcolor{gray!20}0.59 \\
            & \cellcolor{gray!20}\NickName$^{\dag}$                         & \cellcolor{gray!20}0.97 & \cellcolor{gray!20}0.88 & \cellcolor{gray!20}0.52 & \cellcolor{gray!20}0.88 & \cellcolor{gray!20}0.65 & \cellcolor{gray!20}0.62 & \cellcolor{gray!20}0.75 \\
            \bottomrule
        \end{tabular}
    }
    
\end{table}

\begin{table}[H]
    \centering
    \caption{\textbf{VBench \citep{vbench} results across closed-source and open-source video models.}
\NickName\space retains strong text-to-video generation capability, achieving performance comparable to its base model \citep{hunyuanvideo}. 
Moreover, \NickName—which uses a stronger VLM as the conditioning encoder—shows clear improvements in semantic score. 
These results demonstrate that our unified framework not only preserves the generative strengths of the base model but also benefits from enhanced multimodal conditioning through stronger VLMs. ${\dag}$ refer to the methods using LLM rewriter \citep{selfforcing}.
}
    \label{tab:vbench}
    \sisetup{group-digits = false} 

    \resizebox{\linewidth}{!}{
        \begin{tabular}{l l c c c c c c c} 
            \toprule
            \textbf{Type} & \textbf{Name} & \textbf{Total $\uparrow$} & \textbf{\makecell{Quality\\score}} & \textbf{\makecell{Semantic\\score}} & \textbf{\makecell{Aesthetic\\quality}} & \textbf{\makecell{Dynamic \\degree}} & \textbf{\makecell{Object\\class}} & \textbf{\makecell{Overall\\consistency}} \\
            \midrule
            \multirow{5}{*}{ \rotatebox[origin=c]{90}{Close source} }
            & Sora \citep{sora}         & 84.28 & 85.51 & 79.35 & 63.46 & 79.91 & 93.93 & 26.26 \\
            & Veo3 \citep{veo3}         & 85.06 & 85.70 & 82.49 & 63.81 & 72.43 & 93.89 & 27.88 \\
            & Kling1.6 \citep{kling}    & 83.40 & 85.00 & 76.99 & 64.81 & 62.22 & 93.34 & 26.04 \\
            & Jimeng \citep{jimeng}     & 81.97 & 83.29 & 76.69 & 68.80 & 38.43 & 89.62 & 27.10 \\
            & Gen-3 \citep{gen3}        & 82.32 & 84.11 & 75.17 & 63.34 & 60.14 & 87.81 & 26.69 \\
            \midrule
            \multirow{6}{*}{ \rotatebox[origin=c]{90}{Open source} }
            & StepVideo \citep{stepvideot2v}    & 81.83 & 84.46 & 71.28 & 61.23 & 53.06 & 80.56 & 27.12 \\
            & CogVideoX-5B \citep{cogvideox}    & 81.91 & 83.05 & 77.33 & 61.88 & 69.51 & 85.07 & 27.65 \\
            & Wan2.1-14B \citep{wan}            & 83.69 & 85.59 & 76.11 & 66.07 & 65.46 & 86.28 & 25.91 \\
            & HunyuanVideo \citep{hunyuanvideo} & 83.24 & 85.09 & 75.82 & 60.36 & 70.83 & 86.10 & 26.44 \\
            & \cellcolor{gray!20}\NickName                         & \cellcolor{gray!20}82.80 & \cellcolor{gray!20}84.00 & \cellcolor{gray!20}78.01 & \cellcolor{gray!20}65.60 & \cellcolor{gray!20}58.33 & \cellcolor{gray!20}91.10 & \cellcolor{gray!20}26.83 \\
            & \cellcolor{gray!20}\NickName$^{\dag}$                & \cellcolor{gray!20}83.17 & \cellcolor{gray!20}83.69 & \cellcolor{gray!20}81.08 & \cellcolor{gray!20}68.11 & \cellcolor{gray!20}55.56 & \cellcolor{gray!20}91.17 & \cellcolor{gray!20}27.00 \\
            \bottomrule
        \end{tabular}
    }

\end{table}

\subsection{Visual Generation}
\label{sec:Visual Generation}

We initiate our quantitative evaluation by assessing foundational Text-to-Image (T2I) and Text-to-Video (T2V) capabilities using Geneval \citep{geneval} and VBench \citep{vbench}.
These benchmarks assess core visual generation abilities including object composition, color fidelity, semantic alignment, visual quality, and temporal dynamics. 
A primary concern in instruction tuning is catastrophic forgetting, where new tasks degrade original performance.
However, despite Stage 3 allocating only a minor fraction of the data budget to standard T2I/T2V samples, \NickName\space retains performance metrics highly comparable to the HunyuanVideo backbone. 
This empirical evidence confirms that our training strategy effectively mitigates model degradation, preserving the strong pre-trained generative priors while accommodating multimodal customization tasks.

We subsequently evaluate \NickName\space on OpenS2V \citep{opens2v}, a benchmark specifically designed for subject-driven, reference-based video generation—a capability widely absent in standard T2V models.
Here, \NickName\space demonstrates clear advantages over its base model. 
As shown in Table \ref{tab:opens2v}, the model demonstrates well-balanced performance, indicating that \NickName\space 
effectively internalizes the customized, reference-aligned generation abilities introduced during later training stages.

Collectively, these results verify that \NickName\space successfully 
safeguards the robust open-domain capabilities of HunyuanVideo,
while substantially expanding the system toward multimodal, reference-guided, and instruction-driven customization. 
This validates the effectiveness of our unified framework in handling hybrid conditioning without architectural conflict.

\begin{table}[ht]
    \centering
    \caption{\textbf{OpenS2V \cite{opens2v} open-domain results on subject-to-video generation.}
The benchmark evaluates multi-reference video generation across diverse categories, including humans, objects, and face identity consistency. 
\NickName\space even surpasses several strong closed-source systems. 
}
    \label{tab:opens2v}
    \sisetup{group-digits = false} 

    \resizebox{\linewidth}{!}{
        \begin{tabular}{l l c c c c c c c c} 
            \toprule
            \textbf{Type} & \textbf{Name} & \textbf{Total $\uparrow$} & \textbf{Aesthetics} & \textbf{\makecell{Motion\\Smoothness}} & \textbf{\makecell{Motion\\Amplitude}} & \textbf{FaceSim} & \textbf{GmeScore} & \textbf{NexusScore} & \textbf{NaturalScore} \\
            \midrule
            \multirow{3}{*}{ \rotatebox[origin=c]{90}{\makecell{Close\\source}} }
            & Pika2.1 \citep{pika2.1}       & 51.88 & 46.88 & 87.06 & 24.71 & 30.38 & 69.19 & 45.40 & 63.32 \\
            & Vidu2.0 \citep{vidu}          & 51.95 & 41.48 & 90.45 & 13.52 & 35.11 & 67.57 & 43.37 & 65.88 \\
            & Kling1.6 \citep{kling}        & 56.23 & 44.59 & 86.93 & 41.60 & 40.10 & 66.20 & 45.89 & 74.59 \\
            \midrule
            \multirow{5}{*}{ \rotatebox[origin=c]{90}{Open source} }
            & SkyReels-A2 \citep{skyreels}  & 52.25 & 39.41 & 87.93 & \underline{25.60} & 45.95 & 64.54 & \underline{43.75} & 60.32 \\
            & MAGREF \citep{magref}         & 52.51 & 45.02 & 93.17 & 21.81 & 30.83 & \underline{70.47} & 43.04 & 66.90 \\
            & Phantom-14B \citep{phantom}   & 56.77 & \underline{46.39} & \textbf{96.31} & \textbf{33.42} & 51.46 & \textbf{70.65} & 37.43 & \underline{69.35} \\
            & VACE-14B \citep{vace}         & \underline{57.55} & \textbf{47.21} & \underline{94.97} & 15.02 & \textbf{55.09} & 67.27 & \textbf{44.08} & 67.04 \\
            & \cellcolor{gray!20}\NickName                     & \cellcolor{gray!20}\textbf{57.85} & \cellcolor{gray!20}45.92 & \cellcolor{gray!20}94.73 & \cellcolor{gray!20}12.30 & \cellcolor{gray!20}\underline{52.00} & \cellcolor{gray!20}69.69 & \cellcolor{gray!20}42.67 & \cellcolor{gray!20}\textbf{71.99} \\
            \bottomrule
        \end{tabular}
    }

\end{table}


\begin{table}[ht]
    \centering
    \caption{\textbf{Image editing results on the ImgEdit \citep{imgedit} benchmark.}
Despite receiving editing supervision only in stage 3, \NickName\space acquires image-editing capability extremely quickly: after just 1k training steps in stage 3, the model already surpasses most open-source baselines. 
With full stage 3 training, \NickName\space further improves, demonstrating the effectiveness of our progressive training strategy.
}
    \label{tab:imgedit}
    \sisetup{group-digits = false} 

    \resizebox{\linewidth}{!}{
        \begin{tabular}{l l c c c c c c c} 
            \toprule
            \textbf{Type} & \textbf{Name} & \textbf{Average $\uparrow$} & \textbf{Adjust} & \textbf{Remove} & \textbf{Replace} & \textbf{Add} & \textbf{Compose} & \textbf{Action} \\
            \midrule
            \multirow{3}{*}{ \rotatebox[origin=c]{90}{\makecell{Close\\source}} }
            & Gemini2.5 \citep{gemini2dot5}              & 4.30 & 4.48 & 4.39 & 4.24 & 4.30 & 3.88 & 4.61 \\
            & GPT4o \citep{gpt4o}                        & 4.30 & 4.52 & 4.09 & 4.45 & 4.36 & 4.10 & 4.83 \\
            & Seedream4 \citep{seedream4}                & 4.46 & 4.52 & 4.47 & 4.52 & 4.44 & 4.29 & 4.78 \\
            \midrule
            \multirow{7}{*}{ \rotatebox[origin=c]{90}{Open source} }
            & Bagel \citep{bagel}                        & 3.20 & 3.31 & 2.62 & 3.30 & 3.56 & 2.38 & 4.17 \\
            & UniWorld-V1 \citep{uniworld}               & 3.26 & 3.64 & 3.24 & 3.47 & 3.82 & 2.96 & 2.74 \\
            & OmniGen2 \citep{omnigen2}                  & 3.44 & 3.06 & 3.20 & 3.74 & 3.57 & 2.52 & \textbf{4.68} \\
            & Step1x-EditV1.1 \citep{step1xedit}         & 4.01 & 4.17 & 3.73 & \underline{4.11} & \textbf{4.26} & \underline{3.97} & 3.84 \\
            & Flux-Kontext-Dev \citep{flux}              & \underline{4.09} & \textbf{4.28} & \underline{3.85} & \textbf{4.22} & 4.09 & 3.48 & 4.47 \\
            & \cellcolor{gray!20}\NickName\space (only $1k$ step)           & \cellcolor{gray!20}3.82 & \cellcolor{gray!20}3.75 & \cellcolor{gray!20}3.03 & \cellcolor{gray!20}3.71 & \cellcolor{gray!20}4.12 & \cellcolor{gray!20}3.23 & \cellcolor{gray!20}4.40 \\
            & \cellcolor{gray!20}\NickName                                  & \cellcolor{gray!20}\textbf{4.18} & \cellcolor{gray!20}\underline{4.25} & \cellcolor{gray!20}\textbf{4.37} & \cellcolor{gray!20}4.00 & \cellcolor{gray!20}\underline{4.18} & \cellcolor{gray!20}\textbf{4.36} & \cellcolor{gray!20}\underline{4.51} \\
            \bottomrule
        \end{tabular}
    }

\end{table}

\begin{table}[h]
    \centering
    \caption{\textbf{Results on the GEdit \citep{step1xedit} benchmark.}
GEdit measures semantic consistency (SC), perceptual quality (PQ), and an overall score (O) under GPT-based evaluation (denoted by G\_). 
Since our base model \citep{hunyuanvideo} lacks text-rendering capability, we exclude the text-change subtask from evaluation. 
}
    \label{tab:gedit}
    \sisetup{group-digits = false} 

    \resizebox{\linewidth}{!}{
        \begin{tabular}{l l | c c c | c c c | c c c} 
            \toprule
            \textbf{Type} & \textbf{Name} & \multicolumn{3}{c|}{\textbf{Average $\uparrow$}}  & \multicolumn{3}{c|}{\textbf{Subject replace}}  & \multicolumn{3}{c}{\textbf{Style change}} \\
            & & G\_SC & G\_PQ & G\_O & G\_SC & G\_PQ & G\_O & G\_SC & G\_PQ & G\_O \\
            \midrule
            \multirow{3}{*}{ \rotatebox[origin=c]{90}{\makecell{Close\\source}} }
            & Gemini2.5 \citep{gemini2dot5}              & 7.48 & 8.30 & 7.17 & 9.17 & 7.54 & 8.12 & 6.65 & 8.05 & 6.75 \\
            & GPT4o \citep{gpt4o}                        & 8.06 & 7.80 & 7.48 & 8.68 & 7.68 & 8.06 & 8.45 & 6.81 & 7.34 \\
            & Seedream4 \citep{seedream4}                & 8.33 & 8.00 & 7.72 & 9.20 & 7.93 & 8.45 & 8.90 & 6.86 & 7.69 \\
            \midrule
            \multirow{6}{*}{ \rotatebox[origin=c]{90}{Open source} }
            & UniWorld-V1 \citep{uniworld}               & 5.04 & \underline{7.56} & 4.98 & 5.77 & \underline{7.03} & 5.52 & 5.50 & \textbf{6.97} & 5.33 \\
            & OmniGen2 \citep{omnigen2}                  & 6.79 & 6.68 & 6.18 & 7.68 & 6.32 & 6.71 & 7.75 & 5.75 & 6.47 \\
            & Flux-Kontext-Dev \citep{flux}              & 7.23 & 7.28 & 6.53 & 8.10 & \underline{7.03} & 7.01 & 6.80 & 6.03 & 5.91 \\
            & Bagel \citep{bagel}                        & \underline{7.52} & 6.69 & 6.54 & \underline{8.62} & 6.53 & \underline{7.31} & 7.93 & 4.92 & 6.04 \\
            & Step1x-EditV1.1 \citep{step1xedit}         & \textbf{7.60} & 7.29 & \underline{6.87} & \textbf{8.73} & \textbf{7.25} & \textbf{7.80} & \textbf{8.48} & 6.20 & \underline{7.11} \\
            & \cellcolor{gray!20}\NickName                & \cellcolor{gray!20}7.26 & \cellcolor{gray!20}\textbf{7.71} & \cellcolor{gray!20}\textbf{6.88} & \cellcolor{gray!20}8.22 & \cellcolor{gray!20}\underline{7.03} & \cellcolor{gray!20}7.30 & \cellcolor{gray!20}\underline{8.05} & \cellcolor{gray!20}\underline{6.83} & \cellcolor{gray!20}\textbf{7.29} \\
            \bottomrule
        \end{tabular}
    }

\end{table}

\subsection{Visual Editing}
\label{sec:Visual Editing}

We assess the visual editing capabilities of \NickName\space across both spatial (image) and spatio-temporal (video) domains.

\paragraph{Image editing.}
For image editing, we adopt two comprehensive benchmarks: ImgEdit \citep{imgedit} and GEdit \citep{step1xedit}, 
which jointly cover a broad range of editing skills including adjustment, removal, replacement, addition, composition and style change. 
Since our base model lacks text-rendering ability, we exclude the text-change subtask in GEdit. 
Even though HunyuanVideo has no inherent editing capability, our progressive training strategy successfully equips \NickName\space with strong instruction-following behavior. 
Notably, when we evaluate the model after only $1k$ training steps in Stage 3, where editing instructions are introduced for the first time, \NickName\space already surpasses many open-source models on ImgEdit—demonstrating the rapid adaptation enabled by our unified architecture and highlighting the inherent advantages of video models when applied to motion-aware image editing tasks \citep{imontage}.
This rapid improvement demonstrates the effectiveness of our unified architecture and training strategy in enabling fast adaptation and efficient skill transfer from generation tasks to editing tasks.


\paragraph{Video editing.}
To thoroughly evaluate the video editing capabilities of \NickName, we conduct a quantitative comparison on the OpenVE-Bench \citep{openve}, a comprehensive benchmark designed for assessing video editing performance. 
We compare \NickName\space against a wide range of open-source baselines, including VACE-14B \citep{vace}, OmniVideo \citep{omnivideo}, InsViE \citep{insvie}, Lucy-Edit \citep{lucyedit}, ICVE \citep{icve}, Ditto \citep{ditto}, and OpenVE-Edit \citep{openve}.
Following the OpenVE-Bench protocol, we employ two advanced Multimodal LLMs, Gemini 2.5 Pro \citep{gemini2dot5} and Qwen3VL \citep{qwen3vl}, as judges to evaluate the editing quality.
As shown in Table \ref{tab:openve_gemini} and Table \ref{tab:openve_qwen}, \NickName\space significantly outperforms all competing methods.
The consensus across both MLLM evaluators further confirms that \NickName\space achieves precise adherence to editing instructions, consistently surpassing the baselines.

\begin{figure}[h]
    \centering
    \includegraphics[width=\linewidth]{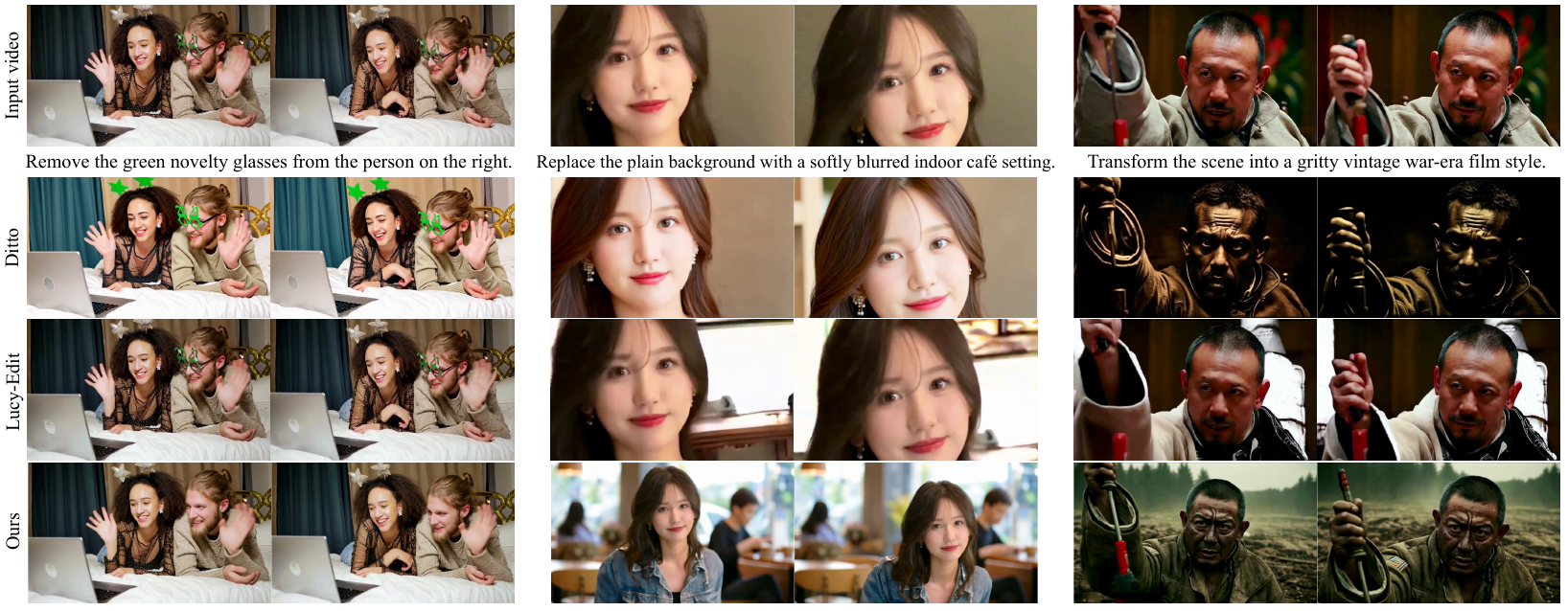}
    \vspace{-2em}
    \caption{\textbf{Qualitative comparison of video editing results between \NickName\space, Ditto \citep{ditto} and Lucy-Edit \citep{lucyedit}.}
Given the same input video and editing instructions, \NickName\space delivers markedly stronger instruction following and better visual quality.
}
    \label{fig:ditto}
\end{figure}

To visually assess the editing capabilities, we compare \NickName\space with Lucy-Edit \citep{lucyedit} and Ditto \citep{ditto}.
Adopting the Ditto evaluation protocol, we utilize their curated mini video editing test set\footnote{\url{https://huggingface.co/datasets/QingyanBai/Ditto-1M/tree/main/mini_test_videos}} and generate diverse editing instructions using GPT-5.
As shown in Figure \ref{fig:ditto}, the comparison highlights that \NickName\space achieves more robust instruction comprehension and higher generative quality than the competing methods.

Together, these results demonstrate that through a unified modeling design and a progressive training pipeline, \NickName\space achieves advanced visual editing capability across both images and videos, offering strong generalization, precise instruction following, and high-quality multimodal editing under a single integrated framework.

\begin{table}[ht]
    \centering
    \caption{\textbf{Quantitative Comparison on OpenVE-Bench \citep{openve} with Gemini 2.5 pro.}
    }
    
    \label{tab:openve_gemini}
    \sisetup{group-digits = false} 
    \resizebox{\linewidth}{!}{
        \begin{tabular}{l | c c c c c c c c c} 
            \toprule
            \textbf{Name} & Overall $\uparrow$ & Global Style & Background Change & Local Change & Local Remove & Local Add & Subtitle Edit & Creative Edit & Camera Edit  \\
            \midrule
            VACE-14B \citep{vace} & 1.57 & 1.49 & 1.55 & 2.07 & 1.46 & 1.26 & 1.48 & 1.47 & 1.62 \\
            OmniVideo \citep{omnivideo} & 1.31 & 1.11 & 1.18 & 1.14 & 1.14 & 1.36 & 1.00 & 2.26 & 1.00 \\
            InsViE \citep{insvie} & 1.53 & 2.20 & 1.06 & 1.48 & 1.36 & 1.17 & 2.18 & 2.02 & 1.09 \\
            Lucy-Edit \citep{lucyedit} & 2.15 & 2.27 & 1.57 & 3.20 & 1.75 & 2.30 & 1.61 & 2.86 & 1.61 \\
            ICVE \citep{icve} & 2.07 & 2.22 & 1.62 & 2.57 & 2.51 & 1.97 & 2.09 & 2.41 & 1.11 \\
            Ditto \citep{ditto} & 1.98 & 4.01 & 1.68 & 2.03 & 1.53 & 1.41 & 2.81 & 1.23 & 1.32  \\
            OpenVE-Edit \citep{openve} & 2.49 & 3.16 & 2.36 & 2.98 & 1.85 & 2.15 & \textbf{2.91} & 2.31 & 2.02 \\
            \rowcolor{gray!20} 
            \NickName                & \textbf{3.18} & \textbf{4.34} & \textbf{2.54} & \textbf{3.73} & \textbf{3.22} & \textbf{2.77} & 2.61 & \textbf{3.29} & \textbf{2.81} \\
            \bottomrule
        \end{tabular}
    }

\end{table}


\begin{table}[ht]
    \centering
    \caption{\textbf{Quantitative Comparison on OpenVE-Bench \citep{openve} with Qwen3VL.}
    }
    
    \label{tab:openve_qwen}
    \sisetup{group-digits = false} 
    \resizebox{\linewidth}{!}{
        \begin{tabular}{l | c c c c c c c c c} 
            \toprule
            \textbf{Name} & Overall $\uparrow$ & Global Style & Background Change & Local Change & Local Remove & Local Add & Subtitle Edit & Creative Edit & Camera Edit  \\
            \midrule
            VACE-14B \citep{vace} & 3.01 & 3.46 & 2.81 & 2.47 & 3.99 & 1.76 & 4.41 & 2.17 & 3.09 \\
            OmniVideo \citep{omnivideo} & 3.66 & 3.41 & 4.11 & 3.75 & \textbf{4.52} & 2.80 & \textbf{4.95} & 1.13 & 3.62 \\
            InsViE \citep{insvie} & 3.25 & 3.63 & 2.68 & 2.82 & 3.56 & 2.25 & 4.77 & 3.36 & 3.61 \\
            Lucy-Edit \citep{lucyedit} & 3.77 & 3.64 & 3.25 & 3.93 & 3.95 & 3.92 & 4.23 & 4.19 & 3.54 \\
            ICVE \citep{icve} & 3.76 & 3.87 & 3.51 & 3.87 & 4.50 & 3.77 & 4.68 & 3.54 & 2.84 \\
            Ditto \citep{ditto} & 3.44 & 4.48 & 3.52 & 2.89 & 3.53 & 2.48 & 3.69 & 4.14 & 3.33  \\
            OpenVE-Edit \citep{openve} & 3.89 & 4.24 & 4.10 & 3.80 & 3.50 & 3.41 & 3.98 & 3.71 & 3.25 \\
            \rowcolor{gray!20} 
            \NickName                & \textbf{4.34} & \textbf{4.78} & \textbf{4.46} & \textbf{4.41} & 4.45 & \textbf{4.43} & 3.39 & \textbf{4.60} & \textbf{4.08} \\
            \bottomrule
        \end{tabular}
    }

\end{table}


\definecolor{mypurple}{RGB}{148, 86, 187}
\definecolor{myblue}{RGB}{11, 164, 234}
\begin{figure}[H]
    \centering
    \includegraphics[width=\linewidth]{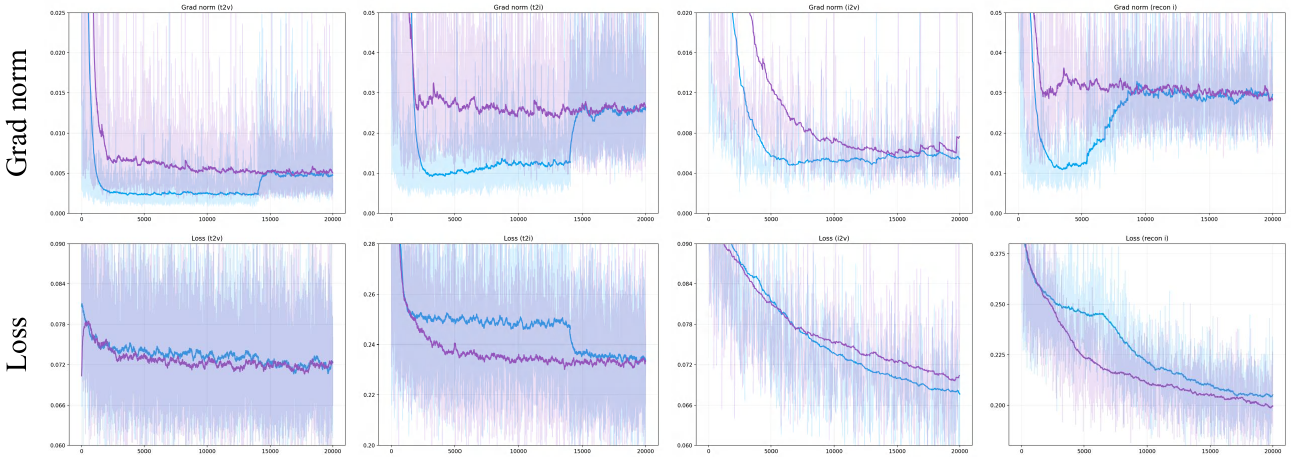}
    \caption{\textbf{Ablation on learnable tokens.}
Training curves comparing models with (\textcolor{mypurple}{purple}) and without (\textcolor{myblue}{blue}) learnable tokens across tasks (T2V, T2I, I2V, and image reconstruction). 
Introducing learnable tokens substantially stabilizes optimization, yielding lower gradient variance, reduced gradient norms, and smoother, healthier loss trajectories. 
This demonstrates that learnable tokens provide more reliable multimodal conditioning and facilitate more stable convergence during unified visual generation training.
}
    \label{fig:curve}
\end{figure}

\begin{figure}[ht]
    \centering
    \includegraphics[width=\linewidth]{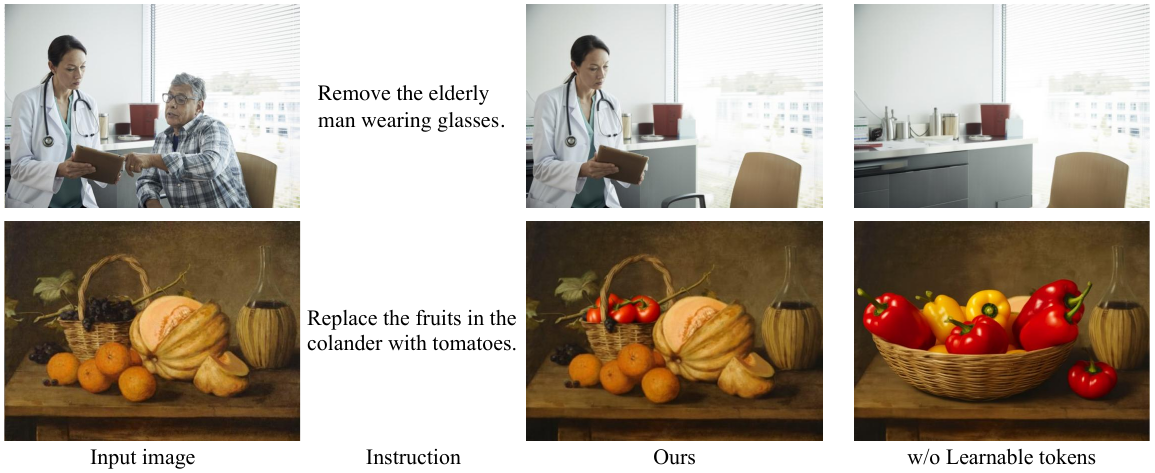}
    \caption{\textbf{Ablation on learnable tokens for image editing.}
Visual comparison of editing results with and without learnable tokens under multimodal conditioning. 
When provided with both the input image and an instruction, the model equipped with learnable tokens produces edits that more faithfully follow the instruction while preserving scene structure and appearance. 
In contrast, the model without learnable tokens often misinterprets the editing intent or generates semantically inconsistent content. 
}
    \label{fig:ablation_mq}
\end{figure}

\begin{figure}[ht]
    \centering
    \includegraphics[width=\linewidth]{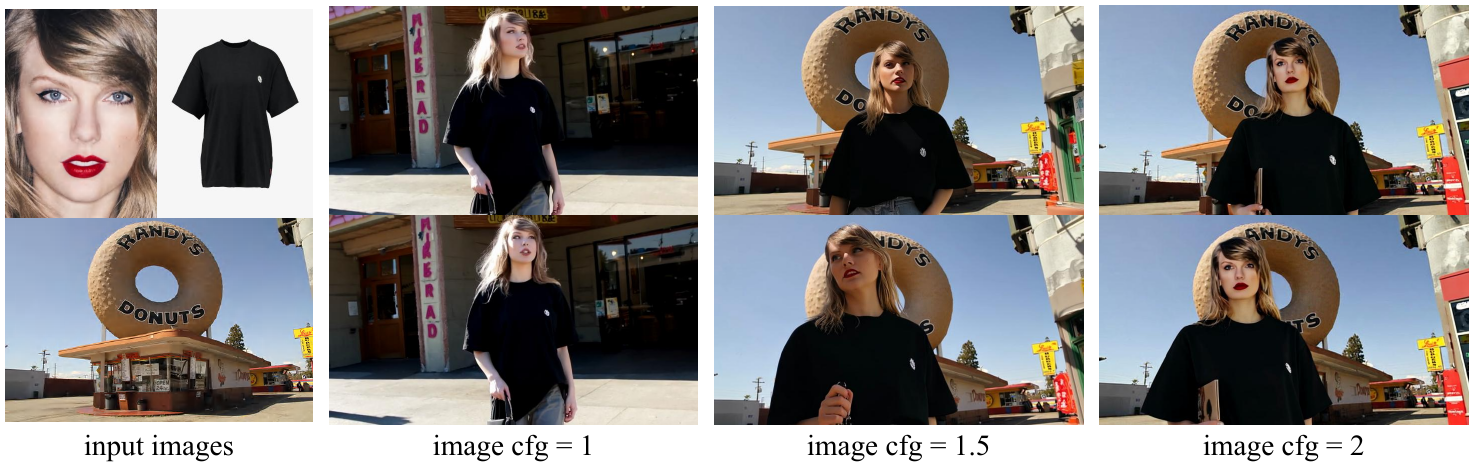}
    \caption{\textbf{Qualitative comparison under different image-CFG scales.}
Increasing the image-CFG weight strengthens adherence to the visual identity provided by the reference images, yielding video frames that more closely match the subject’s appearance and attributes. 
However, excessively large image-CFG values over-constrain the generative process, suppressing motion diversity and leading to noticeably reduced temporal dynamics. 
}
    \label{fig:imgcfg}
\end{figure}

\begin{figure}[ht]
    \centering
    \includegraphics[width=\linewidth]{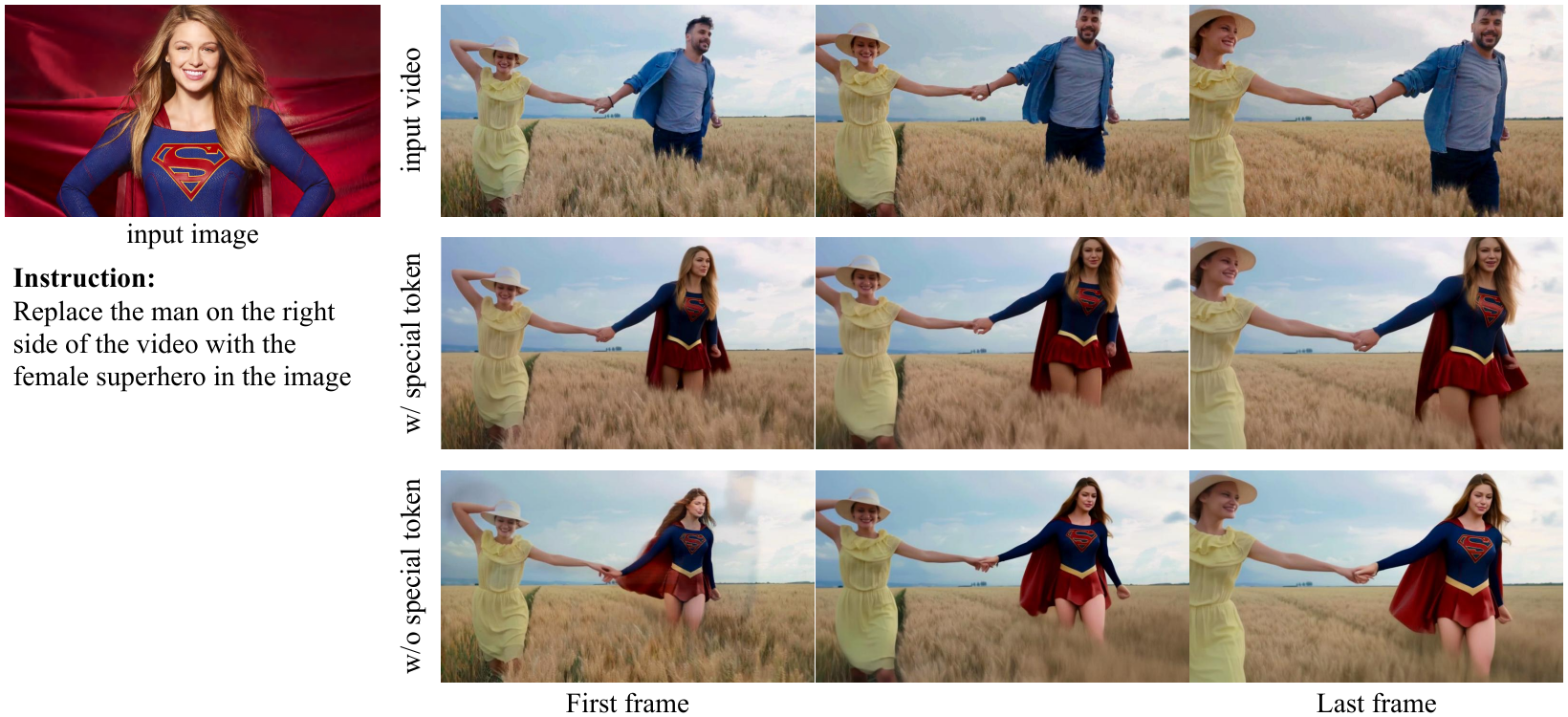}
    \caption{\textbf{Ablation on the special token for separating VAE latents}.
Without special token (bottom row), the model incorrectly entangles the temporal structure of the input video with static image latents, causing noticeable artifacts—most prominently distorted structures in the first generated frame. 
}
    \label{fig:specialtoken}
\end{figure}

\subsection{Ablation Study}
\label{sec:Ablation Study}

We conduct comprehensive ablation studies to isolate and validate the contributions of three pivotal components in our framework:
(1) the integration of learnable tokens for robust multimodal alignment, 
(2) the impact of image classifier-free guidance (Image CFG) on inference dynamics, 
and (3) the special token inserted between VAE latents to disentangle heterogeneous latent sequences.

\paragraph{Learnable tokens.}
As shown in Figure \ref{fig:curve}, 
introducing learnable tokens leads to a substantially more stable training process, characterized by smoother and more well-behaved optimization curves across all tasks (T2V, T2I, I2V, and reconstruction). 
In contrast, removing these tokens results in noticeably noisier gradients and less stable convergence behavior.
Beyond stability, learnable tokens also enhance multimodal conditioning fidelity and semantic precision. 
As illustrated in Figure \ref{fig:ablation_mq}, the model equipped with learnable tokens performs more accurate object removal and replacement, demonstrating that these tokens 
are essential for effectively incorporating multimodal conditions, directly translating into better performance in complex generation and editing tasks.

\paragraph{The guidance of the reference images.}
We investigate the role of Image CFG as a control knob for the fidelity-motion trade-off in reference-guided generation.
Increasing the Image CFG scale significantly reinforces visual adherence, ensuring that the generated results strictly preserves the identity and content of the reference image (Figure \ref{fig:imgcfg}). 
However, we observe a diminishing return: excessively high guidance scales constrain the generative process, suppressing the model's inherent motion priors and causing the output to degenerate into static, frozen sequences. 
Consequently, a moderate image CFG is essential to strike the optimal balance, maintaining high identity fidelity while allowing for natural temporal dynamics.

\paragraph{Special Token Between Latents.}
Finally, we analyze the use of a special token to disentangle VAE latents from different modalities and different lengths. 
Simply concatenating VAE latents from static images and dynamic videos introduces ambiguity, leading the attention mechanism to erroneously interpret the static reference latents as part of the temporal video sequence. 
This misalignment manifests as severe artifacts, particularly structural distortions in the initial frames, as shown in the bottom row of Figure \ref{fig:specialtoken}. 
Furthermore, this explicit separation ensures generalization robustness, allowing the model to handle variable-length inputs during inference without overfitting to specific training sequence patterns.

Together, these ablations confirm that these components are not redundant but foundational: they collectively ensure optimization stability, precise controllability, and effective modality disentanglement in multimodal generation.

\section{Related Works}

Recent advancements in visual generation have coalesced into a dynamic landscape characterized by three pivotal trends. 
First, Diffusion Probabilistic Models have emerged as the de facto paradigm for high-fidelity synthesis and editing, providing the fundamental generative backbone with versatile conditioning interfaces (Section \ref{sec:Diffusion-Based Generation and Editing}). 
Concurrently, mirroring the unification seen in Large Language Models (LLMs), there is a paradigm shift towards Unified and Omni-Visual Generators, where disparate tasks and modalities are consolidated into single, versatile frameworks (Section \ref{sec:Unified and Omni Visual Generation}). 
Complementing these structural evolutions, the maturation of Vision–Language Models (VLMs) has revolutionized generative control, enabling systems to leverage rich semantic understanding for precise guidance and evaluation (Section \ref{sec:Vision–Language Models for Editing and Generative Control}). 
Our work is situated at the intersection of these developments, leveraging their collective strengths to enable robust multimodal conditional generation.

\subsection{Diffusion-Based Generation and Editing}
\label{sec:Diffusion-Based Generation and Editing}

Diffusion models have enabled high-fidelity image \citep{sd3,sd,flux,sana} and video \citep{sanavideo,hunyuanvideo,wan,cogvideox} synthesis from natural language prompts and become the de facto backbone for a wide range of downstream editing methods. 
Instruction-based editing methods \citep{instructpix2pix,imontage} learn to follow free-form text instructions to modify an input image by training on synthetic paired data.
Other lines of work focus on adding explicit spatial control signals: ControlNet \citep{controlnet} augments a frozen text-to-image model with trainable conditional branches,
enabling precise geometric or structural control while preserving the base model’s generation quality.
Subsequent research explores training-free or inversion-based editing, using techniques like null-text inversion  \citep{nulltext}, iterative noising \citep{renoise}, or latent token manipulation \citep{masactrl} to reconstruct an input image in the diffusion latent space and then apply text-guided edits while maintaining identity and background.

\subsection{Unified and Omni Visual Generation}
\label{sec:Unified and Omni Visual Generation}

Inspired by how large language models unify diverse NLP tasks, recent work has started to pursue “unified” visual generators. 
Some works~\citep{omnigen2,uniworld,fulldit} propose diffusion models for unified image generation that support text-to-image, reference-guided generation, image editing, and other visual-conditional tasks within a single framework, largely through a carefully designed conditioning space and multi-task training. 
On the video side, unified video frameworks ~\citep{univid} aim to jointly support video understanding, generation, and instruction-based editing, typically by sharing a video encoder/decoder while exposing different heads or objectives for various tasks. 
There are also works~\citep{vace} on unified in-context video editing that attempt to handle multiple editing conditions in a single model, often by conditioning on example clips or prompts and relying on a generalized editing prior.
A recent survey on unified models for image understanding and generation further emphasizes the trend toward “all-in-one” models, but also points out that most current systems remain limited either to images or to a narrow space of video tasks.

\subsection{Vision–Language Models for Editing and Generative Control}
\label{sec:Vision–Language Models for Editing and Generative Control}

As VLMs have rapidly advanced in perception, reasoning, and instruction following, recent work increasingly leverages them as controllers, teachers, or evaluators for diffusion-based editing. 
Data-centric approaches such as HQ-Edit~\citep{hqedit} use GPT-4V~\citep{gpt4v} and DALL·E 3~\citep{dalle3} to synthesize high-quality editing pairs and automatically score alignment and coherence, effectively turning VLMs into both dataset generators and supervisory signals. 
Model-centric methods~\citep{fireedit,npedit} integrate VLMs directly into the editing pipeline to provide more precise and localized edits or use as a training reward.
On the evaluation side, VLM-based metrics~\citep{hqedit,step1xedit} provide instruction-aware assessments by detecting what changed and whether the modifications adhere to user intent. 
Collectively, these works demonstrate a growing trend in which VLMs play a central role in instruction decomposition, semantic localization, supervision, and evaluation for controllable generative editing.

\section{Conclusion}
\label{sec:Conclusion}

We presented \NickName, a unified visual generator capable of performing both image and video generation and editing under a single framework. 
By carefully designing the model components and a conditioning pipeline that accepts interleaved omnimodal context, \NickName\space can seamlessly integrate heterogeneous inputs and handle a broad spectrum of visual tasks. 
Extensive comparisons demonstrate the effectiveness and strong performance of our approach. 
Moreover, our progressive training strategy enables the model to retain the generative strengths of its base video backbone while acquiring robust multitask capabilities, ultimately yielding a coherent and unified visual generator. 
\NickName\space offers a flexible, scalable foundation for many-to-many visual generation and paves the way toward more general-purpose multimodal generative systems.

\paragraph{Limitations and future works.}
Despite its versatility, \NickName\space inherits several limitations. 

\begin{itemize}
    \item Our base model lacks text-rendering capability, which places \NickName\space at a disadvantage on benchmarks that explicitly evaluate text editing or text generation.

    \item Existing instruction-based editing datasets are generally of lower quality compared with large-scale generation datasets. 
    These editing samples often contain limited motion and simpler visual structures, which may bias the target distribution. 
    Consequently, after incorporating instruction-edit tasks, the model may exhibit slightly reduced visual fidelity or motion richness compared with the original base generator. 
    These limitations highlight the need for higher-quality multimodal editing datasets and more balanced training strategies in future work.

    \item Within MMDiT the complexity of full attention grows quadratically. 
    Therefore, supplying a reference video together with a large number of reference images, the inference latency increases substantially. 
    This suggests that exploring more efficient visual generation backbones is a promising direction.

    \item The modalities currently supported in \NickName\space are ultimately constrained by the VLM. 
    Although certain modalities can be converted into others (e.g., transforming audio into text, or using video to represent motion), our training pipeline only considers the most general modalities—namely text, image, and video. 
    Exploring more powerful and more comprehensive VLMs constitutes a promising direction for future research.
    
\end{itemize}

\newpage
\section*{Acknowledgements}

This work was done during Junyi Chen's internship at Kling Team, Kuaishou Technology.
We acknowledge Google DeepMind for the paper template~\citep{veo3} inspiration.

\bibliographystyle{abbrvnat}
\bibliography{refs}

\newpage
\appendix

\section*{Appendix}

\section{Base Models}
\label{app: Base Models}

\paragraph{Qwen3-VL}
Qwen3-VL \citep{qwen3vl} is an advanced vision–language model belonging to the Qwen3 series. 
It extends large-language-model (LLM) capabilities into vision by integrating image and video encoders, enabling reasoning over diverse visual inputs and natural language instructions. 
This structure allows the model to build cross-modal attention without relying on handcrafted fusion modules, providing strong representational alignment across text and vision modalities. 
Given these properties, Qwen3-VL serves as our visual–instruction foundation: it provides cross-modal alignment, efficient visual tokenization, and empirically strong visual understanding and instruction-following capabilities, which are all essential for many-to-many reference-guided generation in our system.

\paragraph{HunyuanVideo}
HunyuanVideo \citep{hunyuanvideo} is a generative video model based on the Multimodal Diffusion Transformer (MMDiT) architecture \citep{sd3}, which leverages a token-based sequence framework to effectively model video generation.
The MMDiT excels at learning multimodal conditioning, using full attention to directly integrate and condition both visual and multimodal inputs, enhancing the model’s ability to generate coherent and contextually aligned visual content.
These characteristics closely match the requirements of many-to-many generation and editing in our unified setting. 
Consequently, HunyuanVideo offers a suitable video base model, enabling our approach to inherit advanced motion dynamics, coherent cross-modal grounding, and more controllable visual conditioning within a single diffusion framework.

\section{More Details}
\label{app: More Details}

\paragraph{Training details.}
For each task, we define a separate dataloader. 
Before each training step, a task is selected probabilistically, and this selection is broadcasted from rank $0$ to synchronize the tasks across all GPUs. 
This approach prevents idle times on any specific GPU by ensuring that all GPUs process the same task in each step.
We use the default vision preprocessing configuration from Qwen3-VL for all visual inputs, but we limit the total number of video frames fed into Qwen3-VL to a maximum of $8$ frames, with the default frame rate set to $1$ FPS.
During training, we randomly drop conditions \citep{cfg,instructpix2pix}, with the drop probability for different modalities being the same but independent, set to $0.1$ for each. 
Additionally, to maintain load balance, when the number of input images for a task becomes too large, we ensure that all GPUs either drop the condition simultaneously or not at all, preventing a significant discrepancy in the final token sequence lengths.

\paragraph{Proxy tasks.}
To enhance the model's performance, we introduce two proxy tasks during the training process. 
These tasks are designed to help the model better learn and understand visual representations and relationships.
In the stage 1, we employ an \textbf{image reconstruction} proxy task. Specifically, we feed an image into the Vision-Language Model (VLM) while bypassing the Variational Autoencoder (VAE). 
The model is trained to reconstruct the input image. 
This task helps the MMDiT model learn and align the visual features extracted by the VLM, 
strengthening the model’s ability to handle the correspondence between high-level semantics and low-level details.
In the stage 2, we introduce an \textbf{image selection} proxy task. 
When multiple reference images are provided, we can refer to each image by its ID in the instruction. 
To help the model learn the relationship between image IDs and their corresponding visual content, this task involves randomly sampling $k$ images and prompting the model to reconstruct the $i$-th image. 
This allows the model to learn the correspondence between the image ID and the image itself, further improving its ability to handle dynamic image inputs and their relationships.

\paragraph{Inference details.}
Our inference method is based on \citep{cfg,instructpix2pix}. 
For text-conditioned generation tasks, we use a classifier-free guidance (cfg) value $w_{text}$ of $7$, and for multimodal condition-controlled generation tasks, we set $w_{text} = 5$ and $w_{image} = 1.5$.
HunyuanVideo natively supports cfg as an input parameter without the need for multiple inferences, but this requires additional post-training, which we did not adopt.

\section{Quantitative Results}
\label{app: Quantitative results}

\subsection{Visual Understanding}

In Table \ref{tab:understanding}, we present the quantitative results related to the understanding capabilities of our model. 
While our approach primarily focuses on generation, it inherently incorporates components for understanding tasks. leveraging the strengths of our base VLM model, 
Although we did not train Qwen3-VL directly, we utilize it to evaluate various understanding-related metrics, 
which serves as an integral part of \NickName. 
We report on several of these metrics to provide a comprehensive view of the model’s performance in visual understanding.

\begin{table}[ht]
    \centering
    \caption{\textbf{Quantitative results on visual understanding benchmarks.}
    We report comparison of visual understanding performance on standard benchmarks.
    Params denotes the number of model parameters involved when performing visual understanding tasks. Results are reported for both understanding-only models and unified understanding–generation models.
    }
    
    \label{tab:understanding}
    \sisetup{group-digits = false} 

    \resizebox{\linewidth}{!}{
        \begin{tabular}{l l | r |c c c  c c c} 
            \toprule
            \textbf{Type} & \textbf{Name} & Params  & MMMU \citep{mmmu} & MMBench-EN \citep{mmbench} & VideoMME$_{w/o~sub}$ \citep{videomme}  & MVBench \citep{mvbench} & OCRBench \citep{ocrbench} & MathVista$_{mini}$ \citep{mathvista} \\
            \midrule
            \multirow{4}{*}{ \rotatebox[origin=c]{90}{Und. only} }
            & VideoLLaMA3 \citep{videollama3}                & 7B & 48.8 &    - & \underline{66.2} & \textbf{69.7} & 828 & 67.1  \\
            & InternVL3.5 \citep{internvl3dot5}              & 4B & \underline{66.6} &    - & 65.4 & 71.2 & 822 & \textbf{77.1 } \\
            & Qwen2.5-VL \citep{qwen2dot5vl}                 & 7B & 58.6 & 83.5 & 65.1 & \underline{69.6} & \underline{864} & 68.2  \\
            & LLaVA-OneVision1.5 \citep{llavaonevision1dot5} & 4B & 52.7 & \underline{84.2} &    - &    - & 800 & 67.9  \\
            \midrule
            \multirow{7}{*}{ \rotatebox[origin=c]{90}{Und. and Gen.} }
            & show-o2 \citep{showo2}            & 7B & 48.9 & 79.3 & 57.4 & 55.8 & -   & -  \\
            & Janus-Pro \citep{januspro}        & 7B & 41.0 & 79.2 &    - &    - & -   & - \\
            & UniWorld-V1 \citep{uniworld}      & 7B & 58.6 & 83.5 &    - &    - & -   & - \\
            & OmniGen2 \citep{omnigen2}         & 3B & 53.1 & 79.1 &    - &    - & -   & -  \\
            & Bagel \citep{bagel}               & 7B & 55.3 & \textbf{85.0} &    - &    - & -   & 73.1 \\
            & MetaQuery-XL \citep{metaquery}    & 7B & 58.6 & 83.5 & 65.1 & \underline{69.6} & \underline{864} & 68.2  \\
            & \cellcolor{gray!20}\NickName      & \cellcolor{gray!20}4B & \cellcolor{gray!20}\textbf{67.4} & \cellcolor{gray!20}83.9 & \cellcolor{gray!20}\textbf{69.3} & \cellcolor{gray!20}68.9 & \cellcolor{gray!20}\textbf{881} & \cellcolor{gray!20} \underline{73.7} \\
            \bottomrule
        \end{tabular}
    }

\end{table}

\subsection{Visual Generation and Editing}

In this section, we present detailed quantitative results of \NickName\space, as summarized in the table \ref{tab:vbench_detail}, \ref{tab:imgedit_detail} and \ref{tab:gedit_detail}.

\begin{table}[ht]
    \centering
    \caption{\textbf{Quantitative results on Vbench \citep{vbench}.} 
    ${\dag}$ refer to the methods using LLM rewriter \citep{selfforcing}.
    }
    \label{tab:vbench_detail}
    \sisetup{group-digits = false}

    \resizebox{\linewidth}{!}{
        \begin{tabular}{l c c c c c c c c c c c c c c c c c c c} 
            \toprule
            \textbf{Name} & \textbf{Total $\uparrow$} & 
            \rotatebox[origin=c]{90}{\textbf{\makecell{Quality\\score}}} &  
            \rotatebox[origin=c]{90}{\textbf{\makecell{Semantic\\score}}} &  
            \rotatebox[origin=c]{90}{\textbf{\makecell{Subject\\consistency}}} &  
            \rotatebox[origin=c]{90}{\textbf{\makecell{Overall \\consistency}}} & 
            \rotatebox[origin=c]{90}{\textbf{\makecell{Temporal\\style}}} &  
            \rotatebox[origin=c]{90}{\textbf{\makecell{Appearance\\style}}} &
            \rotatebox[origin=c]{90}{\textbf{\makecell{Scene}}} &  
            \rotatebox[origin=c]{90}{\textbf{\makecell{Spatial \\relationship}}} & 
            \rotatebox[origin=c]{90}{\textbf{\makecell{Color}}} &  
            \rotatebox[origin=c]{90}{\textbf{\makecell{Human\\action}}} &
            \rotatebox[origin=c]{90}{\textbf{\makecell{Multiple\\objects}}} &  
            \rotatebox[origin=c]{90}{\textbf{\makecell{Object \\class}}} & 
            \rotatebox[origin=c]{90}{\textbf{\makecell{Image\\quality}}} &  
            \rotatebox[origin=c]{90}{\textbf{\makecell{Aesthetic\\quality}}} &
            \rotatebox[origin=c]{90}{\textbf{\makecell{Dynamic\\degree}}} &  
            \rotatebox[origin=c]{90}{\textbf{\makecell{Motion \\Smoothness}}} & 
            \rotatebox[origin=c]{90}{\textbf{\makecell{Temporal\\flickering}}} &  
            \rotatebox[origin=c]{90}{\textbf{\makecell{Background\\consistency}}} \\
            \midrule
            \NickName          & 82.80 & 84.00 & 78.01 & 95.99 & 26.83 & 26.29 & 24.73 & 46.74 & 67.85 & 90.77 & 94.00 & 69.04 & 91.09 & 64.43 & 65.58 & 58.33 & 98.73 & 99.31 & 97.65 \\
            \NickName$^{\dag}$ & 83.17 & 83.69 & 81.08 & 95.87 & 26.99 & 24.73 & 24.96 & 51.00 & 75.85 & 90.22 & 97.80 &
            83.16 & 91.17 & 62.23 & 68.11 & 55.56 & 98.72 & 99.04 & 97.69 \\
            \bottomrule
        \end{tabular}
    }

\end{table}

\begin{table}[ht]
    \centering
    \caption{\textbf{Quantitative results on ImgEdit \citep{imgedit}.} 
    }
    \label{tab:imgedit_detail}
    \sisetup{group-digits = false} 
    
    \resizebox{\linewidth}{!}{
        \begin{tabular}{c c c c c c c c c c} 
            \toprule
             \textbf{Average} & adjust & style & background & extract & remove & replace & add & compose & action \\
            \midrule
            4.18 & 4.25 & 4.44 & 3.99 & 3.56 & 4.37 & 4.00 & 4.18 & 4.36 & 4.51 \\
            \bottomrule
        \end{tabular}
    }

\end{table}

\begin{table}[ht]
    \centering
    \caption{\textbf{Quantitative results on GEdit \citep{step1xedit}.} 
    }
    \label{tab:gedit_detail}
    \sisetup{group-digits = false} 
    
    \resizebox{\linewidth}{!}{
        \begin{tabular}{c c c | c c c | c c c | c c c | c c c | c c c | c c c | c c c | c c c | c c c | c c c} 
            \toprule
              \multicolumn{3}{c|}{\textbf{Average $\uparrow$}}  
            & \multicolumn{3}{c|}{\textbf{background change}}  
            & \multicolumn{3}{c|}{\textbf{color}}  
            & \multicolumn{3}{c|}{\textbf{material}}  
            & \multicolumn{3}{c|}{\textbf{motion change}}  
            & \multicolumn{3}{c|}{\textbf{ps human}}  
            & \multicolumn{3}{c|}{\textbf{style change}}  
            & \multicolumn{3}{c|}{\textbf{subject add}}  
            & \multicolumn{3}{c|}{\textbf{subject remove}}  
            & \multicolumn{3}{c|}{\textbf{subject replace}}  
            & \multicolumn{3}{c}{\textbf{tone transfer}} \\
            G\_SC & G\_PQ & G\_O & 
            G\_SC & G\_PQ & G\_O & 
            G\_SC & G\_PQ & G\_O & 
            G\_SC & G\_PQ & G\_O & 
            G\_SC & G\_PQ & G\_O & 
            G\_SC & G\_PQ & G\_O & 
            G\_SC & G\_PQ & G\_O & 
            G\_SC & G\_PQ & G\_O & 
            G\_SC & G\_PQ & G\_O & 
            G\_SC & G\_PQ & G\_O & 
            G\_SC & G\_PQ & G\_O \\
            \midrule
            7.26& 7.71& 6.88 &
            7.90& 7.48& 7.30 &
            8.40& 7.53& 7.51 &
            7.55& 7.03& 7.10 &
            4.83& 8.58& 4.91 &
            4.41& 8.51& 4.90 &
            8.05& 6.83& 7.29 &
            8.17& 7.82& 7.73 &
            7.98& 8.26& 7.63 &
            8.22& 7.03& 7.30 &
            7.05& 8.00& 7.09 \\
            \bottomrule
        \end{tabular}
    }

\end{table}



\section{Qualitative results}
\label{app: Qualitative results}

\begin{figure}[ht]
    \centering
    \includegraphics[width=\linewidth]{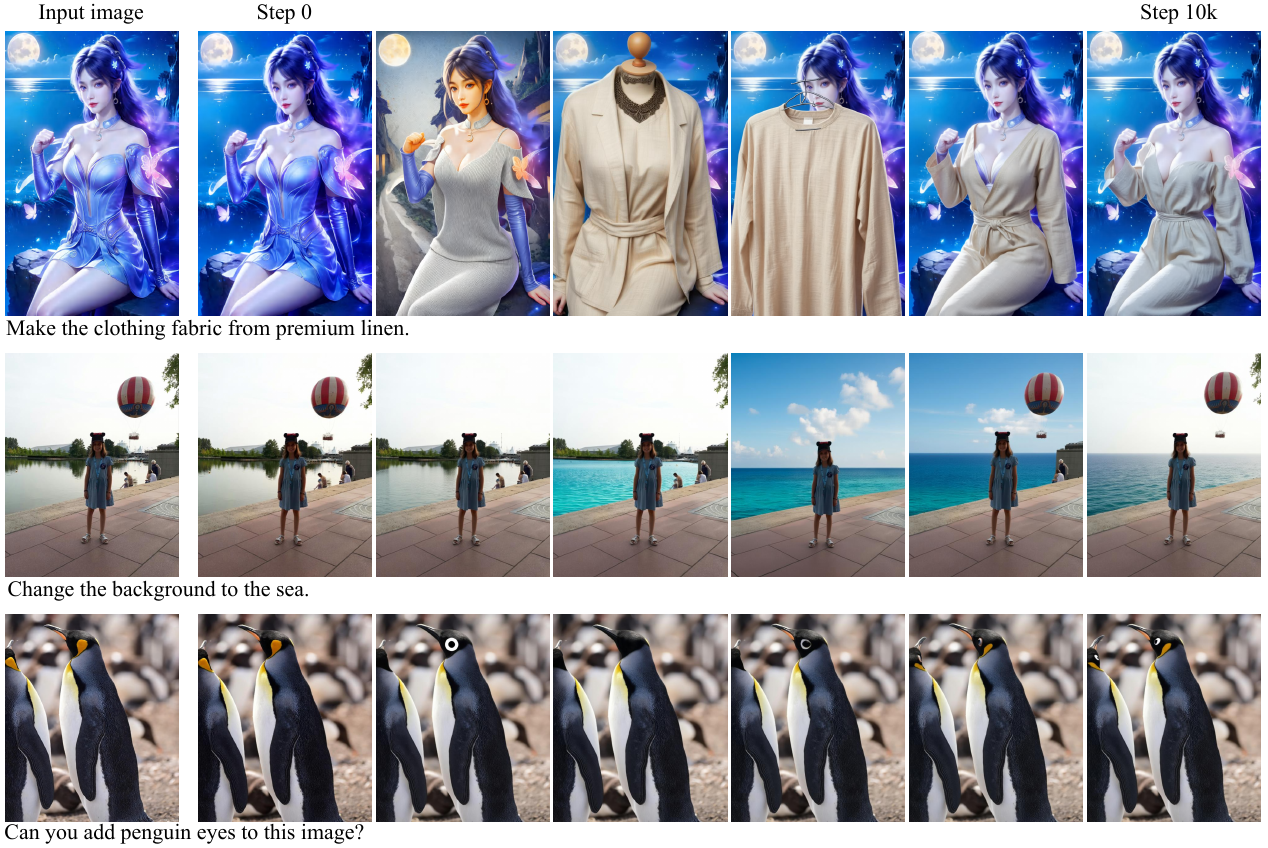}
    \caption{\textbf{Training progression from Stage 3, demonstrated through image editing tasks.} 
    The model benefits from the strong initialization achieved in earlier stages, gradually improving its ability to align visuals with instructions. As training progresses, the model refines its understanding, capturing finer details and showing clear progress in handling different tasks.}
    \label{fig:edit_process}
\end{figure}

\begin{figure}[ht]
    \centering
    \includegraphics[width=\linewidth]{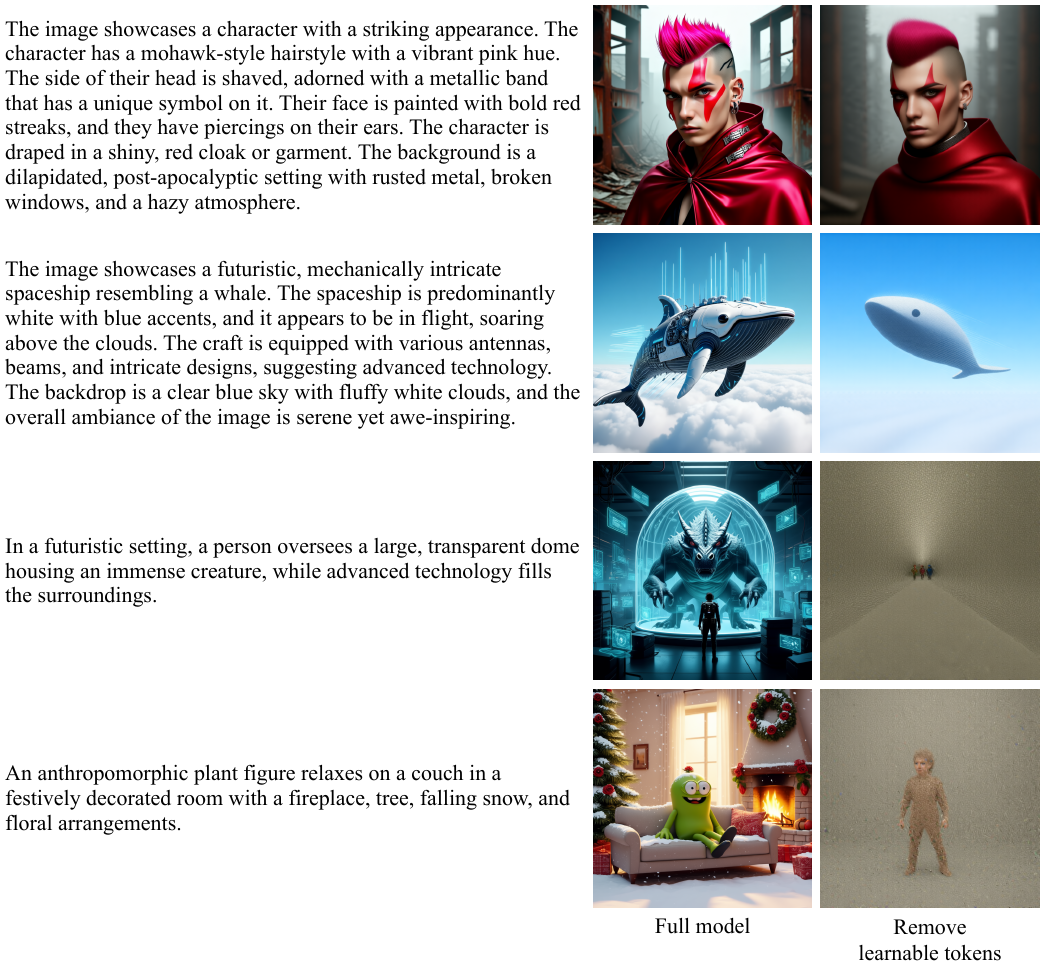}
    \caption{The effect of learnable tokens in the final model. Use the same initial gaussian noise.}
    \label{fig:remove_mq}
\end{figure}

\paragraph{Stage 3 training process.}
As shown in Figure \ref{fig:edit_process}, in Stage 3, the training process benefits greatly from the alignment achieved in the previous two stages, providing a strong initialization. 
At the start of training, the model begins by learning the correspondence between visuals and instructions. 
As training continues, it gradually improves and starts to capture more detailed features. 
Over time, the model becomes better at handling more complex details, showing clear progress as the training steps increase.

\paragraph{The effect of learnable tokens.}
In Figure \ref{fig:remove_mq}, we demonstrate the role of the learnable token in the final model. 
When we choose longer captions, the model allocates more attention to the text features that are related to high-level semantics, while the learnable token learns to represent more low-level details. 
This aligns with our initial design: the learnable token acts as a more flexible interface for the VLM, optimized together with MMDiT, allowing the model to more effectively learn knowledge at different levels. 
However, when shorter captions are selected, the role of the learnable token becomes more pronounced.

\end{document}